\documentclass{article}

\usepackage{microtype}
\usepackage{graphicx}
\usepackage{subfigure}
\usepackage{booktabs} 
\usepackage{dblfloatfix}
\usepackage{placeins}
\usepackage{multirow}
\usepackage{algorithm}
\usepackage{algorithmic}

\usepackage{hyperref}

\usepackage[accepted]{icml2026}

\usepackage{amsmath}
\usepackage{amssymb}
\usepackage{mathtools}
\usepackage{amsthm}

\usepackage[capitalize,noabbrev]{cleveref}

\theoremstyle{plain}

\theoremstyle{definition}

\theoremstyle{remark}

\icmltitlerunning{Lagrangian Perturbation Diffusion Steering}

\begin{document}

\twocolumn[

\icmltitle{Lagrangian Perturbation Diffusion Steering: Latent Reinforcement Learning for Generative Policies}

\begin{icmlauthorlist}
\icmlauthor{Hikmet Simsir}{bilkent}
\icmlauthor{Ozgur S. Oguz}{bilkent}
\end{icmlauthorlist}

\icmlaffiliation{bilkent}{Department of Computer Engineering, Bilkent University, Ankara, Türkiye}

\icmlcorrespondingauthor{Hikmet Simsir}{hikmet.simsir@bilkent.edu.tr}

\icmlkeywords{Reinforcement Learning, Diffusion Models, Generative Policies, Policy Fine-tuning, Constrained Optimization, Trust Region Methods, Offline-to-Online RL, Robotic Manipulation}

\vskip 0.3in
]
\printAffiliationsAndNotice{}
\begin{abstract}

Behavior cloning with high-capacity generative policies achieves strong imitation performance, but is often limited by demonstration coverage and distribution shift. Direct reinforcement learning fine-tuning can improve performance, but updating large action decoders is frequently unstable and sample inefficient.  We propose \textbf{Lagrangian Perturbation Diffusion Steering (LP-DS)}, a lightweight adaptation method that improves a frozen generative policy by learning a compact noise-space perturbation before decoding. LP-DS optimizes this perturbation with a Lagrangian trust-region objective, improving downstream value while constraining deviation from the latent prior. Across RoboMimic manipulation, OpenAI Gym locomotion, and Adroit dexterous manipulation benchmarks, LP-DS improves sample efficiency, success, and return while maintaining higher action-space entropy than unconstrained noise-space steering, with return improvements of up to 25\% over prior baselines. Additional evaluations with flow-matching backbones, a large vision-language-action model, and physical Franka deployment show that LP-DS is not limited to compact diffusion policies or simulated benchmarks. Project page: \url{https://sites.google.com/view/lp-ds/home}.

\end{abstract}

\section{Introduction}

\begin{figure}[t!]
    \centering
    \includegraphics[width=\columnwidth]{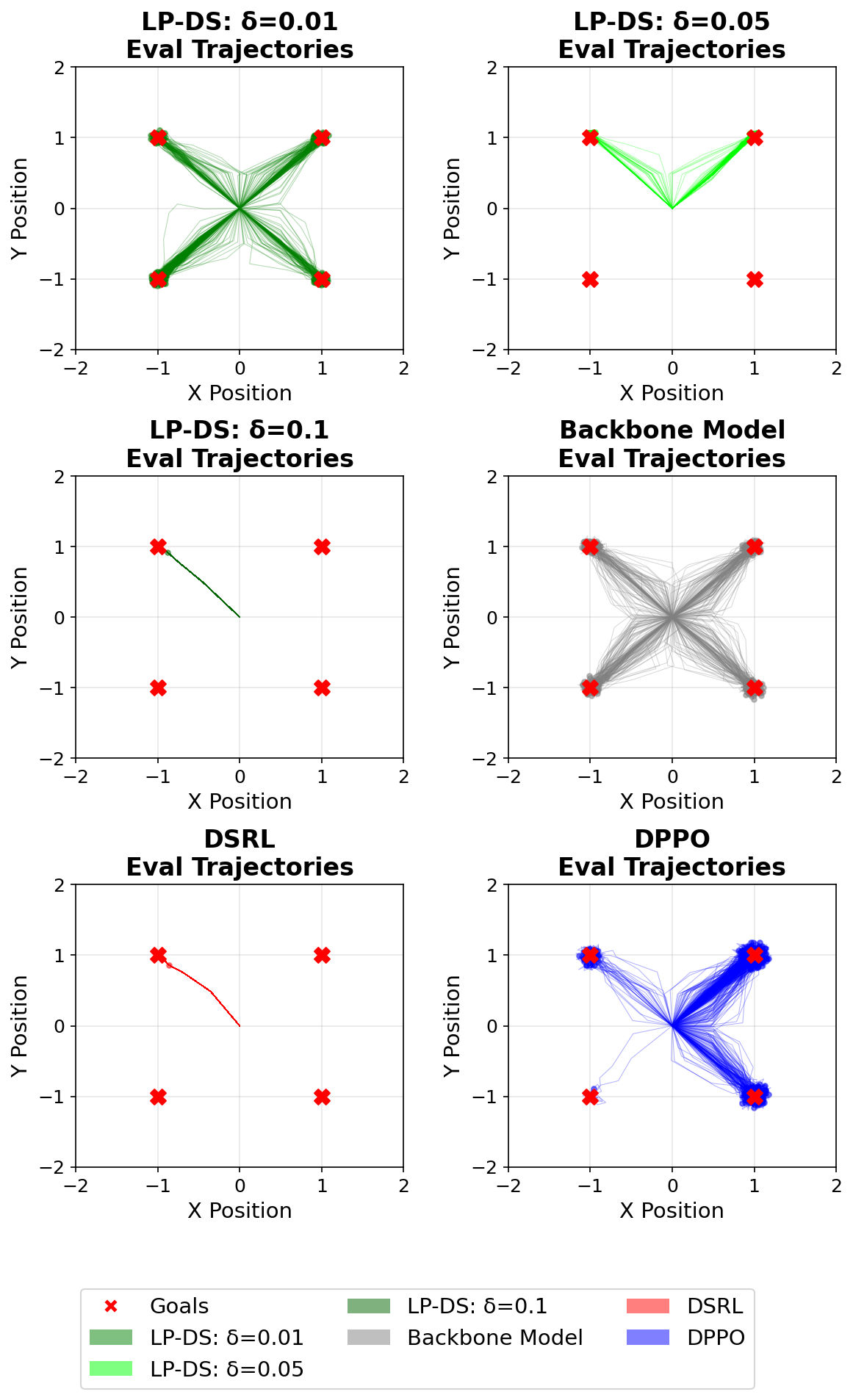}
    \caption{\textbf{Toy multi-goal navigation with symmetric rewards.}
    Four equally optimal Gaussian reward peaks (red markers) define four target modes.
    We visualize evaluation rollouts from the frozen backbone and after adaptation using LP-DS with different trust-region bounds $\delta \in \{0.01, 0.05, 0.1\}$, alongside DSRL \citep{wagenmaker2025steeringdiffusionpolicylatent} and DPPO \citep{ren2024diffusionpolicypolicyoptimization}. See Sec.~\ref{sec:toy_delta} for detailed analysis.}
    \label{fig:toy_multimodal_collapse}
\end{figure}

High-capacity generative policies, including diffusion \citep{wang2023diffusionpoliciesexpressivepolicy} and flow-matching decoders \citep{black2024pi0, park2025flowqlearning}, have emerged as a powerful paradigm for continuous control and robotics because they can represent expressive, multimodal action distributions learned from offline data \citep{ho2022imagenvideohighdefinition}. Despite strong imitation performance, behavior cloning (BC) policies often inherit limitations of the dataset: demonstrations provide incomplete state-space coverage, omit rare recovery behaviors, and can lead to brittle performance under distribution shift \citep{mehta2024stablebc}. Reinforcement learning (RL) \citep{sutton2018reinforcement} offers a natural mechanism for online improvement, but directly fine-tuning large generative decoders can be unstable and sample intensive, especially when optimization interacts with long-horizon denoising or integration dynamics \citep{chandra2025diwadiffusionpolicyadaptation}.

A promising alternative is to improve a frozen generative policy by acting in its latent noise space, treating the decoder as a black-box map from noise to actions. Recent work on \emph{Diffusion Steering via Reinforcement Learning} (DSRL) shows that latent-space RL can efficiently steer diffusion policies without modifying the decoder \citep{wagenmaker2025steeringdiffusionpolicylatent}. However, DSRL trains a latent policy that effectively replaces the pre-training prior. This can cause two failure modes: the steered noise can drift into low-density regions of the standard Gaussian prior used to train the decoder, and the learned latent policy can collapse the backbone's multimodal behavior into a narrow subset of modes, as illustrated in Figure~\ref{fig:toy_multimodal_collapse}.

We propose \textbf{Lagrangian Perturbation Diffusion Steering (LP-DS)}, a lightweight method for improving frozen generative policies by learning a state- and noise-conditioned residual in latent noise space. LP-DS shifts Gaussian noise inputs via $w = \epsilon + \Delta_\theta(s,\epsilon)$ and optimizes $\Delta_\theta$ with a \emph{Lagrangian trust-region objective} that improves downstream value while limiting deviation from the latent prior. This keeps latent queries closer to high-density regions of the pretrained latent prior while still allowing task-directed policy improvement.

\begin{figure}[t!]
    \centering
    \includegraphics[width=\columnwidth]{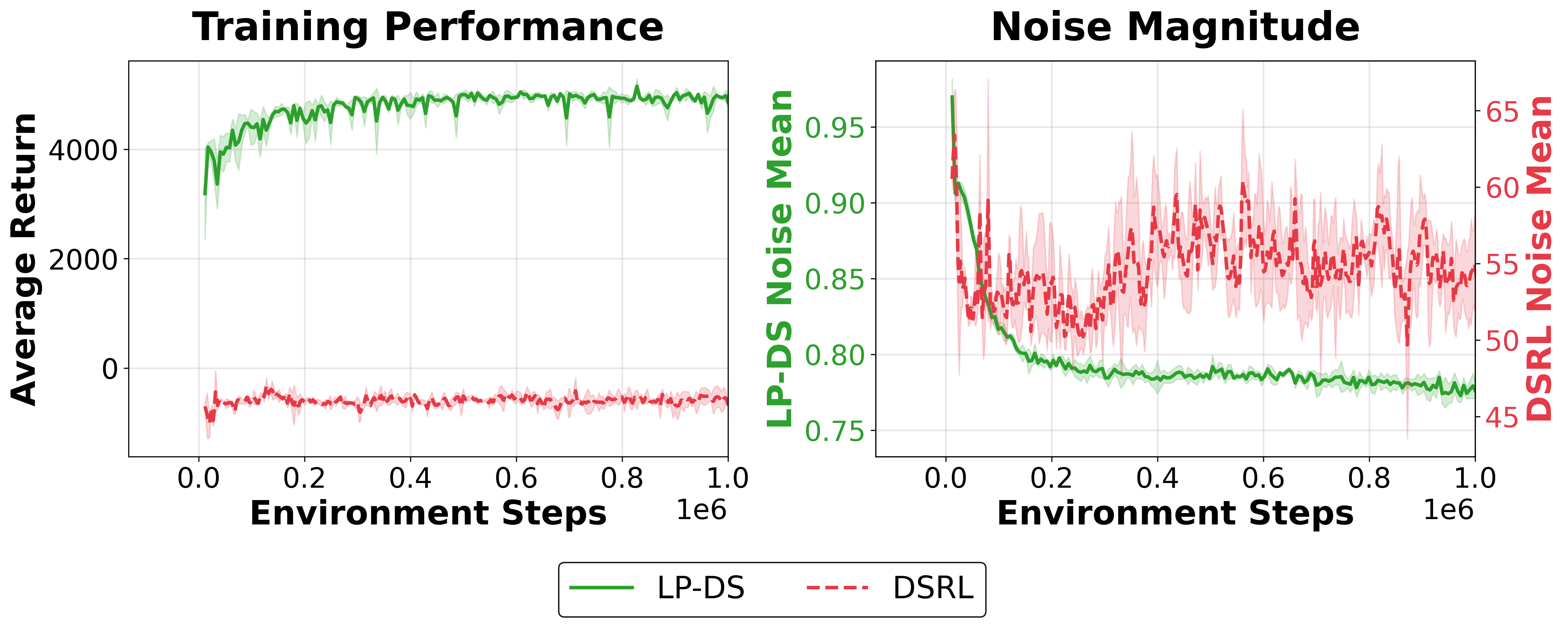}
    \caption{\textbf{Failure mode of weakly constrained noise-space steering.} Success rate and latent magnitude during online adaptation on \textsc{HalfCheetah-v2} \citep{brockman2016openaigym}, averaged over 3 seeds and using the configuration of Section~\ref{sec:experiments} under the same hard clip ($\lVert w \rVert \le 100$). DSRL predicts higher-magnitude latent queries that correlate with low-prior-density decoder inputs and performance degradation, while LP-DS stays closer to high-density prior regions by design.}
    \label{fig:noise_magnitude}
\end{figure}

Figure~\ref{fig:noise_magnitude} makes this failure mode observable: under weakly constrained noise-space steering, DSRL predicts high-magnitude latent queries that correlate with unstable decoding and performance degradation, whereas LP-DS maintains smaller perturbations through its adaptive trust-region constraint. This distinction is central to LP-DS: rather than replacing the generative prior with an unconstrained latent policy, it learns a constrained residual shift around the prior.

We evaluate LP-DS across diverse continuous-control and robotics domains and find consistent improvements over representative noise-space steering and diffusion fine-tuning baselines. LP-DS improves success and return while better preserving the behavioral diversity of the frozen backbone, as measured by action-space entropy. We further validate LP-DS across backbone classes, including matched diffusion/flow-matching comparisons, a large vision-language-action $\pi_0$ backbone, and physical Franka robot experiments with simulation-based latent adaptation.

Our main contributions are:
\begin{itemize}
\item We introduce \textbf{Lagrangian Perturbation Diffusion Steering (LP-DS)}, a lightweight adaptation framework that improves frozen generative policies by learning a state- and noise-conditioned residual in latent noise space, avoiding the instability and cost of full decoder fine-tuning.
\item We propose a \textbf{Lagrangian trust-region objective} that dynamically constrains the perturbation magnitude. The trust-region target $\delta$ provides a controllable mechanism for balancing reward maximization against preservation of the pretrained latent prior.
\item We show that LP-DS mitigates mode collapse relative to prior-replacing noise-space steering, preserving higher action-space diversity while achieving strong performance across manipulation, locomotion, and dexterous-control benchmarks.
\item We validate LP-DS beyond compact diffusion policies through matched diffusion/flow-matching comparisons, a large vision-language-action $\pi_0$ backbone, and physical Franka robot experiments with simulation-based latent adaptation.
\end{itemize}
\section{Related Work}
Behavior cloning (BC) with high-capacity generative policies has become a strong paradigm for visuomotor control, particularly through diffusion-based policies \citep{chi2024diffusionpolicyvisuomotorpolicy,pearce2023imitatinghumanbehaviourdiffusion}. Recent work has further improved their generalization, data efficiency, and scalability across robotic settings \citep{ze20243ddiffusionpolicygeneralizable,sridhar2023nomadgoalmaskeddiffusion,dasari2024ingredientsroboticdiffusiontransformers,nvidia2025gr00tn1openfoundation}.

Reinforcement learning (RL) provides principled tools for improving policies beyond demonstration performance, and offline RL in particular has become a major setting for leveraging large static datasets without requiring additional environment interaction \citep{kostrikov2021offlinereinforcementlearningimplicit}. A growing line of work studies the intersection of offline RL and generative policy classes, where expressive decoders can model complex, multimodal action distributions while value-based objectives guide policy improvement \citep{wang2023diffusionpoliciesexpressivepolicy, chen2023offlinereinforcementlearninghighfidelity, kang2023efficientdiffusionpoliciesoffline}. Beyond standard RL formulations, conditional generative modeling has also been shown to function as a competitive decision-making paradigm on offline benchmarks by directly generating actions conditioned on returns or constraints \citep{ajay2023conditionalgenerativemodelingneed}. In robotics, recent work further emphasizes data efficiency in imitation and policy learning pipelines, highlighting complementary directions that reduce demonstration requirements for complex tasks \citep{ankile2024juicerdataefficientimitationlearning}. More recently, diffusion policies have been integrated into actor--critic style offline RL via implicit Q-learning, yielding diffusion-parameterized policies that can be improved with stable value learning \citep{hansenestruch2023idqlimplicitqlearningactorcritic}. In parallel, online fine-tuning methods that differentiate through diffusion policy rollouts further connect generative policies with policy-gradient optimization \citep{ren2024diffusionpolicypolicyoptimization}, highlighting both the promise and the practical challenges of adapting high-capacity generative backbones with RL objectives.

A related theme across generative modeling and decision making is to perform optimization directly in the \emph{noise} or \emph{latent} space of a pretrained model, leveraging the strong behavioral prior induced by the generative backbone. In vision, several works improve diffusion outputs by modifying or optimizing the initial noise distribution at inference time, including reward-driven noise optimization for one-step generators \citep{eyring2024renoenhancingonesteptexttoimage}, noise-dependent initialization and pruning effects in denoising \citep{mao2024lotterytickethypothesisdenoising}, and latent-space strategies for generating rare concepts from pretrained diffusion models \citep{samuel2023generatingimagesrareconcepts}. More broadly, latent-space inference and sampling procedures have been studied as a computationally efficient way to approximate posterior inference and control generation \citep{venkatraman2025outsourceddiffusionsamplingefficient}, while in RL, pretrained behavioral priors similarly serve as strong scaffolds for downstream policy improvement \citep{singh2020parrotdatadrivenbehavioralpriors}. Closest to our setting, latent-space reinforcement learning has recently been used to steer diffusion policies by learning a noise-space policy that replaces the base prior during action generation \citep{wagenmaker2025steeringdiffusionpolicylatent}. In contrast, LP-DS keeps the pretrained prior as an explicit reference distribution and learns a \emph{residual} state- and noise-conditioned perturbation with an adaptive Lagrangian trust-region constraint, directly targeting stable online improvement while limiting off-prior latent queries. Finally, recent amortization approaches learn auxiliary networks to predict beneficial noise modifications and reduce test-time compute for diffusion generation \citep{eyring2025noisehypernetworksamortizingtesttime}; unlike these methods, our perturbation module is optimized \emph{with online RL signals} for control and is explicitly regularized to preserve the multimodal structure of a frozen policy backbone rather than accelerating or approximating a fixed generation objective. Broadly, this paradigm of guiding a frozen generative prior with a learned value function parallels high-level planning frameworks like SayCan~\citep{ahn2022icanisay}, adapted here to the continuous latent space of low-level control.
\section{Preliminaries}
\label{sec:preliminaries}

\paragraph{Markov decision processes.}
We model decision making as a Markov decision process (MDP) $\mathcal{M}=(\mathcal{S},\mathcal{A},P,r,\gamma)$, where $\mathcal{S}$ and $\mathcal{A}$ denote the state and action spaces, $P(s' \mid s,a)$ is the transition kernel, $r(s,a)$ is the reward function, and $\gamma\in[0,1)$ is the discount factor.
A policy $\pi(a\mid s)$ induces trajectories via $a_t\sim\pi(\cdot\mid s_t)$ and $s_{t+1}\sim P(\cdot\mid s_t,a_t)$.
The reinforcement learning objective is to maximize the expected discounted return
$J(\pi)=\mathbb{E}_{\pi,P}\!\left[\sum_{t=0}^{\infty}\gamma^t r(s_t,a_t)\right]$.

The \emph{state–action value function} (Q-function) of a policy $\pi$ is defined as
\[
Q^\pi(s,a)
:= \mathbb{E}_{\pi,P}\!\left[
\sum_{t=0}^{\infty}\gamma^t r(s_t,a_t)
\;\middle|\;
s_0=s,\; a_0=a
\right],
\]
i.e., the expected discounted return obtained by taking action $a$ in state $s$ and subsequently following policy $\pi$.
The corresponding state value function is $V^\pi(s)=\mathbb{E}_{a\sim\pi(\cdot\mid s)}[Q^\pi(s,a)]$.

\paragraph{Diffusion models.}
Diffusion models define a latent-variable generative process via a \emph{forward} noising Markov chain and a learned \emph{reverse} denoising chain \citep{ho2020denoising}.
The forward process incrementally perturbs data with Gaussian noise:

\begin{equation}
\begin{aligned}
x_t = \sqrt{1-\beta_t}\,x_{t-1} + \sqrt{\beta_t}\,\epsilon_t, \\
\qquad
\epsilon_t \sim \mathcal{N}(0,I),
\quad t=1,\dots,T .
\label{eq:forward_step}
\end{aligned}
\end{equation}

This implies the closed-form marginal $x_t = \sqrt{\bar{\alpha}_t}\,x_0 + \sqrt{1-\bar{\alpha}_t}\,\epsilon$ with $\epsilon\sim\mathcal{N}(0,I)$ and $\bar{\alpha}_t=\prod_{i=1}^t(1-\beta_i)$.
Generation runs the reverse-time chain from $x_T\sim\mathcal{N}(0,I)$ using a learned transition kernel
\begin{equation}
\begin{aligned}
p_\theta(x_{t-1}\mid x_t)
&:= \mathcal{N}\!\big(
x_{t-1};
\mu_\theta(x_t,t),
\Sigma_t
\big), \\
&\qquad t = T, \dots, 1 .
\end{aligned}
\label{eq:reverse_kernel}
\end{equation}
Here $\mu_\theta$ is produced by a neural network and $\Sigma_t$ is typically fixed by the noise schedule \citep{ho2020denoising}.
Denoising Diffusion Implicit Models (DDIM) introduces an implicit (optionally deterministic) reverse update that enables fewer sampling steps while preserving the same training objective \citep{song2020denoising}.

\paragraph{Flow matching.}
Flow matching provides a general ODE-based generative framework by learning a velocity field $v_\theta$ that transports a base noise distribution $p_0$ to the data distribution $p_1$ \citep{lipman2023flowmatching,liu2022rectified}.
The model defines an ODE $dx/dt = v_\theta(x,t)$ for $t\in[0,1]$, integrated from noise $x(0)$ to data $x(1)$.
Training minimizes a regression objective matching $v_\theta$ to the velocity of a conditional probability path $x_t = \psi_t(x_{\mathrm{n}}, x_{\mathrm{d}})$:
\begin{align}
\min_\theta\;
\mathbb{E}_{t, x_{\mathrm{n}}, x_{\mathrm{d}}}\Big[
\big\|v_\theta(x_t,t) - \dot{\psi}_t(x_{\mathrm{n}}, x_{\mathrm{d}})\big\|_2^2
\Big],
\end{align}
where $x_{\mathrm{n}}\sim \mathcal{N}(0, I)$, $x_{\mathrm{d}}$ is a data sample, and $\dot{\psi}_t$ is the time derivative of the interpolation. Sampling is performed by numerically integrating the ODE from $t=0$ to $t=1$.

\paragraph{Problem setup.}
We consider generative policies that predict action chunks using diffusion or flow-matching backbones \citep{chi2024diffusionpolicyvisuomotorpolicy,park2025flowqlearning}. Under deterministic sampling schemes (e.g., DDIM), generation acts as a deterministic mapping from initial noise to output actions \citep{venkatraman2025outsourceddiffusionsamplingefficient}. We therefore model the frozen backbone as a black-box decoder $\Phi: \mathcal{S} \times \mathcal{W} \rightarrow \mathcal{A}$, where $\mathcal{W} = \mathbb{R}^d$ denotes the latent noise space \citep{wagenmaker2025steeringdiffusionpolicylatent}. Our objective is to improve performance in the MDP $\mathcal{M}$ by adapting the latent input $w$ without updating the potentially heavy decoder $\Phi$.

\section{Methodology}
\label{sec:method}
We present \textbf{Lagrangian Perturbation Diffusion Steering (LP-DS)}, a framework for adapting frozen generative policies via residual noise control. We formalize policy improvement as a constrained optimization problem: we seek to maximize downstream value by perturbing the latent input of the generative backbone, subject to a trust-region constraint that keeps the modified noise distribution close to the backbone's training prior.

\subsection{Steering via Residual Perturbation}
\label{sec:residual_perturb}

Let $\Phi:\mathcal{S}\times\mathcal{W}\rightarrow\mathcal{A}$ be a frozen, deterministic generative decoder (e.g., a diffusion or flow-matching model) that maps a latent noise vector $w \in \mathcal{W}$ and state $s$ to an action chunk. In the standard setting, actions are sampled by drawing $w$ from a fixed prior, typically $\epsilon \sim \mathcal{N}(0, I)$.

To steer the policy toward higher-reward behaviors without modifying the decoder weights, we parameterize the input noise as a state- and noise-conditioned residual transformation of the base prior. We define a learnable perturbation network $\Delta_\theta: \mathcal{S} \times \mathcal{W} \rightarrow \mathcal{W}$ that also depends on the base noise $\epsilon$, and generate latent queries according to:
\begin{equation}
w = \epsilon + \Delta_\theta(s, \epsilon), \quad \text{where } \epsilon \sim \mathcal{N}(0, I).
\label{eq:residual_noise}
\end{equation}
This formulation induces a steered policy $\pi_\theta(a|s)$ defined by the pushforward of the base noise distribution through the residual mapping and the frozen decoder $\Phi$. When $\Delta_\theta(s, \epsilon) = 0$, the policy recovers the original behavior cloning distribution exactly.

\begin{algorithm}[t]
\caption{Lagrangian Perturbation Diffusion Steering}
\label{alg:lp_ds}
\begin{algorithmic}[1]
\STATE \textbf{Init:} $\Delta_\theta(\cdot) \approx 0$, Critics $Q^{\mathcal{A}}_\psi, Q^{\mathcal{W}}_\phi$, $\alpha \ge 0$, Buffer $\mathcal{B}$
\FOR{each env step}
    \STATE \textbf{// 1. Data Collection}
    \STATE Sample $\epsilon \sim \mathcal{N}(0, I)$; \quad Set $w \leftarrow \epsilon + \Delta_\theta(s, \epsilon)$
    \STATE $a \leftarrow \Phi(w; s)$; Step env; Store $(s, a, r, s')$ in $\mathcal{B}$
    
    \STATE \textbf{// 2. Action Critic Update (TD Error)}
    \STATE Sample batch $B \sim \mathcal{B}$

    \STATE \textit{// Compute target using next-state learned policy}
    \STATE Sample $\epsilon' \sim \mathcal{N}(0, I)$; \quad $w' = \epsilon' + \Delta_{\theta'}(s', \epsilon')$
    
    \STATE $a' = \Phi(w'; s')$; \quad $y = r + \gamma \bar{Q}^{\mathcal{A}}(s', a')$
    \STATE $\mathcal{L}_{\psi} = \mathbb{E}_B [(Q^{\mathcal{A}}_\psi(s, a) - y)^2]$; \quad Update $\psi$ using $\nabla \mathcal{L}_{\psi}$
    
    \STATE \textbf{// 3. Latent Critic Distillation}
    \STATE \textit{// Distill on base noise distribution}
    \STATE $\mathcal{L}_{\phi} = \mathbb{E}_{s \sim B, \epsilon} [(Q^{\mathcal{W}}_\phi(s, \epsilon) - Q^{\mathcal{A}}_\psi(s, \Phi(\epsilon; s)))^2]$
    \STATE Update $\phi$ using $\nabla \mathcal{L}_{\phi}$
    
    \STATE \textbf{// 4. Actor \& Dual Update (Lagrangian)}
    \STATE $\mathcal{L}_{\theta} = \mathbb{E}_{s \sim B, \epsilon} \big[ Q^{\mathcal{W}}_\phi(s, \epsilon+\Delta_\theta(s, \epsilon)) - \alpha(\|\Delta_\theta(s, \epsilon)\|^2 - \delta) \big]$
    \STATE $\theta \leftarrow \theta + \eta_\theta \nabla_\theta \mathcal{L}_{\theta}$
    \STATE $\alpha \leftarrow [\alpha + \eta_\alpha \mathbb{E}_{B, \epsilon} (\|\Delta_\theta(s, \epsilon)\|^2 - \delta)]_+$
    
    \STATE Update target networks
\ENDFOR
\end{algorithmic}
\end{algorithm}


\subsection{Constrained Optimization via Lagrangian Relaxation}
\label{sec:lagrangian_optimization}

Our primary objective is to find parameters $\theta$ that maximize the expected return of the steered policy. However, unconstrained optimization in the latent space allows $\Delta_\theta(s,\epsilon)$ to grow arbitrarily large, pushing $w$ into low-density regions of the standard Gaussian prior. Such low-probability latent queries can fall far outside the regions typically encountered during backbone training, which leads to erratic action generation and mode collapse.

We therefore formulate the learning problem as maximizing the expected Q-value of the generated actions subject to a trust-region constraint. For a given state $s$, let $q_\theta(\cdot \mid s)$ denote the conditional distribution of the steered latent $w = \epsilon + \Delta_\theta(s,\epsilon)$ induced by the base noise $\epsilon \sim p_0 = \mathcal{N}(0, I)$. To keep the steered noise distribution close to the prior on average over encountered states, we constrain the \emph{expected conditional} Kullback--Leibler (KL) divergence between $q_\theta(\cdot \mid s)$ and the prior $p_0$:
\begin{equation}
\mathbb{E}_{s \sim \mathcal{D}}\!\left[ D_{\mathrm{KL}}\!\left( q_\theta(\cdot \mid s)\,\|\,p_0 \right) \right].
\label{eq:expected_cond_kl}
\end{equation}

Following \citet{eyring2025noisehypernetworksamortizingtesttime}, under the regularity conditions of the residual transformation and in the small-Jacobian regime, we approximate this expected conditional KL divergence by the expected squared perturbation magnitude:
\begin{equation}
\label{eq:kl_approx}
\begin{split}
\mathbb{E}_{s \sim \mathcal{D}}\!\left[ D_{\mathrm{KL}}\!\left( q_\theta(\cdot \mid s)\,\|\,p_0 \right) \right] \\
\approx \frac{1}{2} \mathbb{E}_{s \sim \mathcal{D}, \epsilon \sim p_0} \left[ \|\Delta_\theta(s, \epsilon)\|_2^2 \right].
\end{split}
\end{equation}

Based on this approximation, we impose a constraint on the expected perturbation magnitude:
\begin{equation}
\begin{aligned}
\max_\theta \quad & \mathbb{E}_{s \sim \mathcal{D}, \epsilon \sim p_0} \left[ Q^{\mathcal{W}}(s, \epsilon + \Delta_\theta(s, \epsilon)) \right] \\
\textrm{s.t.} \quad & \mathbb{E}_{s \sim \mathcal{D}, \epsilon \sim p_0} \left[ \|\Delta_\theta(s, \epsilon)\|_2^2 \right] \le \delta,
\end{aligned}
\label{eq:constrained_opt}
\end{equation}
where $Q^{\mathcal{W}}$ is a critic estimating the value of latent noise vectors and $\delta > 0$ is a hyperparameter controlling the allowable deviation from the prior.

We solve this constrained problem via the method of Lagrange multipliers. We introduce a dual variable (Lagrange multiplier) $\alpha \ge 0$ and define the Lagrangian $\mathcal{L}(\theta, \alpha)$:
\begin{equation}
\mathcal{L}(\theta, \alpha) = \mathbb{E} \left[ Q^{\mathcal{W}}(s, w) - \alpha \left( \|\Delta_\theta(s, \epsilon)\|_2^2 - \delta \right) \right].
\label{eq:lagrangian_primal_dual}
\end{equation}

We then seek the saddle point of the min-max objective:
\begin{equation}
\min_{\alpha \ge 0} \max_\theta \mathcal{L}(\theta, \alpha).
\end{equation}
This formulation naturally creates an adaptive regularization mechanism. If the perturbation violates the trust region ($\mathbb{E}_{s,\epsilon}[\|\Delta_\theta(s,\epsilon)\|_2^2] > \delta$), $\alpha$ increases, forcing the actor to prioritize the prior constraint over reward maximization. Conversely, if the perturbation is small, $\alpha$ decreases, allowing the actor to explore more aggressive steering.

We alternate between updating the actor $\theta$ and the multiplier $\alpha$. The actor is updated via gradient ascent on Equation~\ref{eq:lagrangian_primal_dual}. The multiplier $\alpha$ is updated via a projected dual update to enforce the constraint $\alpha \ge 0$:
\begin{equation}
\alpha_{k+1} \leftarrow \left[ \alpha_k + \eta_\alpha \mathbb{E}_{s, \epsilon} \left[ \|\Delta_\theta(s, \epsilon)\|_2^2 - \delta \right] \right]_+.
\end{equation}

\begin{figure*}[t]
    \centering
    \includegraphics[width=\textwidth]{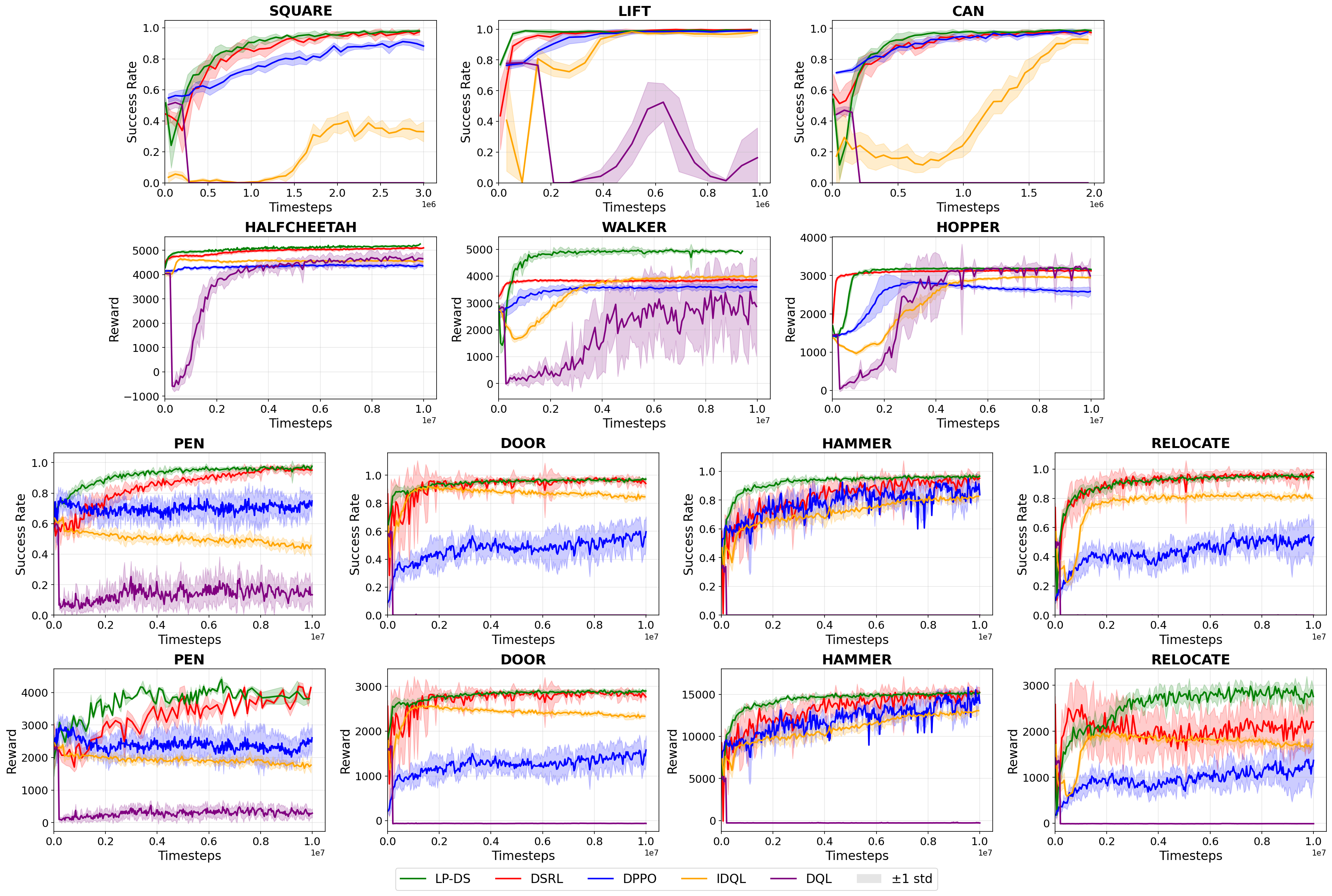}
    \caption{\textbf{Baseline comparisons across domains.}
    \textbf{Top row:} RoboMimic manipulation success rates.
    \textbf{Second row:} OpenAI Gym locomotion episodic returns.
    \textbf{Third row:} Adroit dexterous manipulation success rates.
    \textbf{Bottom row:} Adroit dexterous manipulation episodic returns.
    We compare LP-DS against DSRL \citep{wagenmaker2025steeringdiffusionpolicylatent}, DPPO \citep{ren2024diffusionpolicypolicyoptimization}, IDQL \citep{hansenestruch2023idqlimplicitqlearningactorcritic}, and DQL \citep{wang2023diffusionpoliciesexpressivepolicy}.
    Across domains, LP-DS improves performance over baselines.}
    \label{fig:baseline_comparison_all}
\end{figure*}


\subsection{Algorithm Summary}
The online training procedure is summarized in Algorithm~\ref{alg:lp_ds}. At each step, we collect experience using the steered policy $w = \epsilon + \Delta_\theta(s,\epsilon)$ (Sec.~\ref{sec:residual_perturb}), update the critics $Q^{\mathcal{A}}$ and $Q^{\mathcal{W}}$ via TD learning and distillation respectively, and finally perform the primal-dual update for $\theta$ and $\alpha$ to maximize reward subject to the trust-region constraint (Sec.~\ref{sec:lagrangian_optimization}).


\section{Experiments}
\label{sec:experiments}

\textbf{Setup.}
We run all methods with the same environment interaction budget and report mean performance across multiple random seeds with $\pm 1$ standard deviation. We evaluate LP-DS on RoboMimic manipulation tasks~\citep{robomimic2021}, 
OpenAI Gym locomotion environments~\citep{brockman2016openaigym}, 
and Adroit dexterous manipulation benchmarks~\citep{rajeswaran2017learning}. We compare \textbf{LP-DS} to three representative categories of generative policy improvement: (i) latent noise-space steering (\textbf{DSRL}), (ii) policy-gradient fine-tuning (\textbf{DPPO}), and (iii) offline-to-online diffusion Q-learning (\textbf{IDQL}, \textbf{DQL}). Throughout, we treat the generative backbone policy as a frozen module for LP-DS and DSRL. For these cases, we optimize only the adaptation mechanism. We refer the reader to Appendix~\ref{app:delta_values} for details on the trust-region target $\delta$ (which is set to $0.35$ in most experiments) and experiments.

\textbf{Backbone Architectures.}
Backbone initialization differs by benchmark family. For RoboMimic and Gym, we use the pre-trained diffusion policies released with the DSRL baseline. For Adroit, we train a \textbf{flow-matching} generative policy from scratch and use it as the backbone for the steering methods (LP-DS and DSRL), leveraging its efficient deterministic ODE-based sampling. This design choice explicitly demonstrates that LP-DS is \emph{not specific to diffusion models}: the proposed perturbation-and-trust-region framework applies to any generative policy that admits a latent-to-action mapping, including continuous-time flow models. For the fine-tuning baselines (DPPO, IDQL, DQL), we use diffusion policy backbones matched in parameter count and initial performance to ensure fair comparison.

\subsection{Benchmark Results}
Figure~\ref{fig:baseline_comparison_all} reports results across RoboMimic, OpenAI Gym, and Adroit benchmarks averaged over 6 random seeds. Overall, \textbf{LP-DS outperforms competing methods on the majority of environments} in terms of success rate or episodic return, while achieving higher sample efficiency and more stable learning dynamics. Across RoboMimic manipulation, LP-DS rapidly reaches high success, especially on precision-sensitive tasks such as \textsc{Square}. In Gym locomotion, LP-DS achieves strong returns with low variance. On WALKER2D, LP-DS reaches approximately 5000 episodic returns compared to roughly 4000 for the strongest baseline, corresponding to a $\sim$25\% improvement. In Adroit dexterous manipulation, LP-DS obtains the strongest overall performance across success and reward metrics.

These results suggest that constrained latent steering prevents off-prior adaptation while still allowing effective online improvement. Compared with DSRL, LP-DS avoids uncontrolled latent drift by keeping perturbations close to the prior. Compared with DPPO and diffusion Q-learning baselines, LP-DS updates only a lightweight perturbation module rather than fine-tuning the full decoder, improving sample efficiency and stability.

\subsection{Diversity and Mode Collapse Under Online Adaptation}
\label{sec:diversity_collapse}

We analyze how different adaptation strategies affect behavioral diversity during online learning.
We study collapse at two complementary levels:
(i) \emph{trajectory-level mode coverage} in a controlled toy domain, and
(ii) \emph{action-space diversity} in high-dimensional benchmarks using a nonparametric entropy metric.

\subsubsection{Toy Domain: Trust-Region Size Controls Mode Collapse}
\label{sec:toy_delta}

Figure~\ref{fig:toy_multimodal_collapse} provides a controlled study designed to isolate \emph{mode collapse} and to illustrate the role of the trust-region parameter $\delta$ in LP-DS.
The environment contains four Gaussian reward peaks placed symmetrically at equal distance from the origin, yielding four equally optimal target modes.
The frozen backbone policy exhibits multi-goal behavior, reaching all modes across evaluation rollouts. See Appendix \ref{sec:toy_example} for further explanation and results.

LP-DS exposes a clear and interpretable trade-off between performance and diversity through the choice of $\delta$.
With a small trust-region bound ($\delta = 0.01$), LP-DS preserves the multimodal structure of the backbone and consistently reaches multiple goals, albeit with less tightly concentrated trajectories.
Increasing the bound to $\delta = 0.05$ yields more directed and higher-quality trajectories while still maintaining coverage of multiple modes.
However, when $\delta$ is further increased ($\delta = 0.1$), LP-DS converges to a single dominant mode, exhibiting behavior consistent with mode collapse despite the symmetric reward landscape.

These results demonstrate that $\delta$ acts as an explicit \emph{tuning parameter} controlling the trade-off between maximizing task performance and preserving behavioral diversity.
Larger values of $\delta$ allow more aggressive latent-space optimization and faster convergence to high-reward trajectories, but at the cost of collapsing onto a smaller subset of modes.
Conversely, smaller values of $\delta$ anchor the policy more strongly to the pretrained prior, preserving multimodality while limiting how sharply the policy can specialize.

For comparison, DSRL collapses immediately to a single mode, while DPPO converges to a subset of modes over training.
In contrast, LP-DS makes this trade-off \emph{explicit and controllable}, enabling users to balance success rate against mode collapse through a single, interpretable parameter.


\subsubsection{The Avoiding Environment: Long-Horizon Multi-Path Diversity}
\label{sec:avoiding_diversity}

To complement the symmetric multi-goal toy domain, we evaluate LP-DS in the \textsc{Avoiding} environment from the diverse-behavior imitation benchmark of \citet{jia2024diversebehaviorsbenchmarkimitation}, designed to test long-horizon trajectory-level multimodality.
In this task, the agent must reach the goal while navigating around obstacles, and multiple qualitatively distinct routes can solve the task. This setting allows us to study whether online adaptation preserves diverse trajectory modes rather than only maintaining local action-space entropy.

\begin{figure}[t]
    \centering
    \includegraphics[width=\columnwidth]{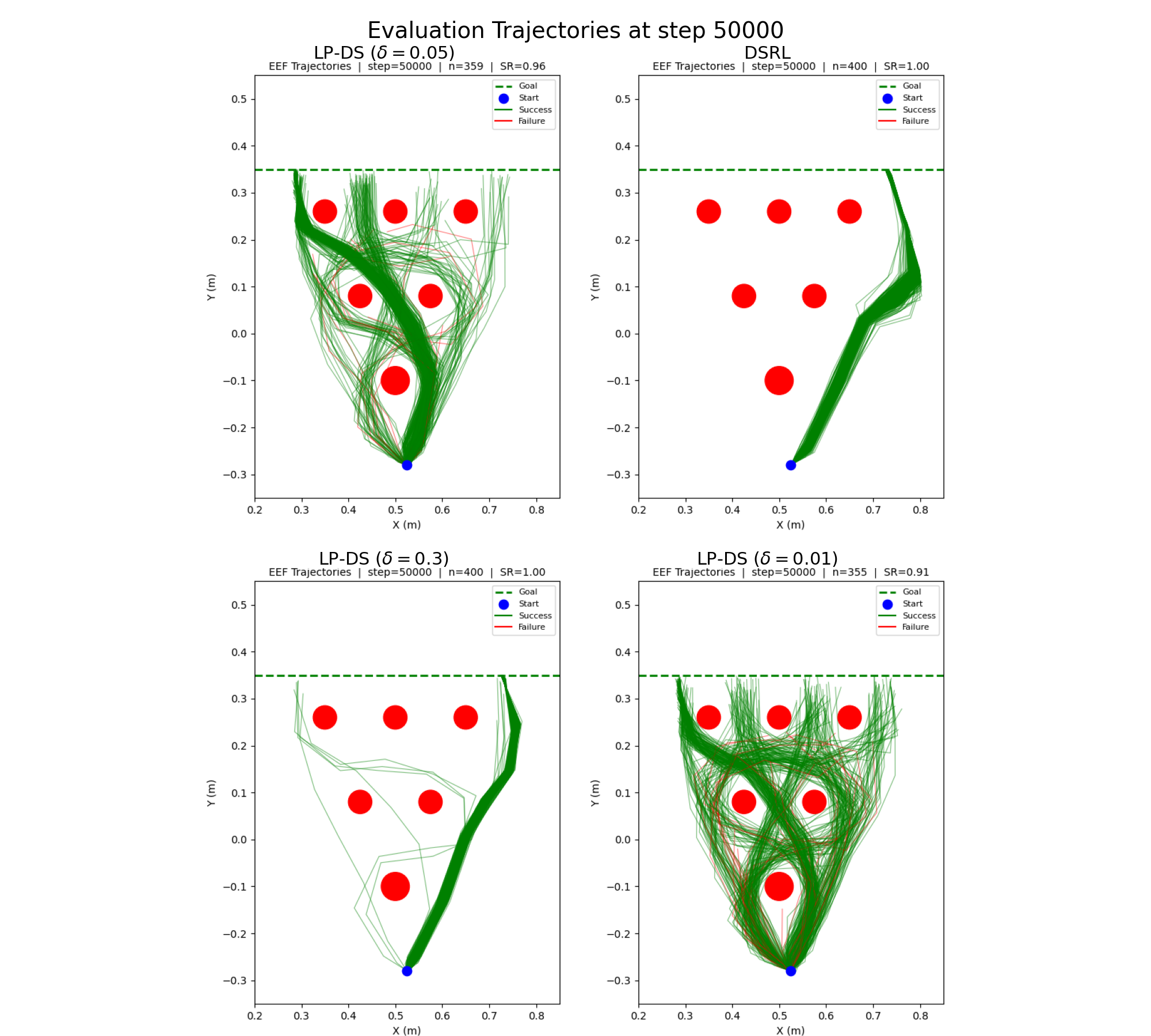}
    \caption{\textbf{Trajectory-level multimodality in the \textsc{Avoiding} environment.}
    We visualize evaluation trajectories at step 50{,}000 for LP-DS with different trust-region targets and for DSRL. Smaller trust-region targets preserve a broader set of feasible obstacle-avoidance routes, while larger targets produce more concentrated, specialized trajectories. DSRL collapses to a single dominant path.}
    \label{fig:avoiding_trajectories}
\end{figure}

Figure~\ref{fig:avoiding_trajectories} visualizes evaluation trajectories at step 50{,}000 for LP-DS with different trust-region targets and for DSRL. The results show that the trust-region target $\delta$ again acts as a controllable specialization--diversity dial. With a small trust-region bound ($\delta=0.01$), LP-DS preserves a broad set of feasible obstacle-avoidance routes, indicating strong trajectory-level multimodality. Increasing the bound to $\delta=0.05$ yields more directed behavior while still maintaining multiple route choices. With a larger bound ($\delta=0.3$), the policy concentrates around a narrower successful route, reflecting stronger task specialization. In contrast, DSRL collapses to a single dominant path.

These results reinforce the conclusion from the toy-domain study: LP-DS provides explicit control over the trade-off between multimodal behavior and reward maximization in a longer-horizon sequential task. Quantitative success-rate and reward curves for the \textsc{Avoiding} environment are provided in Appendix~\ref{app:avoiding}.

\subsubsection{Action-Space Diversity During Online Adaptation}
\label{sec:entropy_online}

\begin{figure}[t]
    \centering
    \includegraphics[width=\columnwidth]{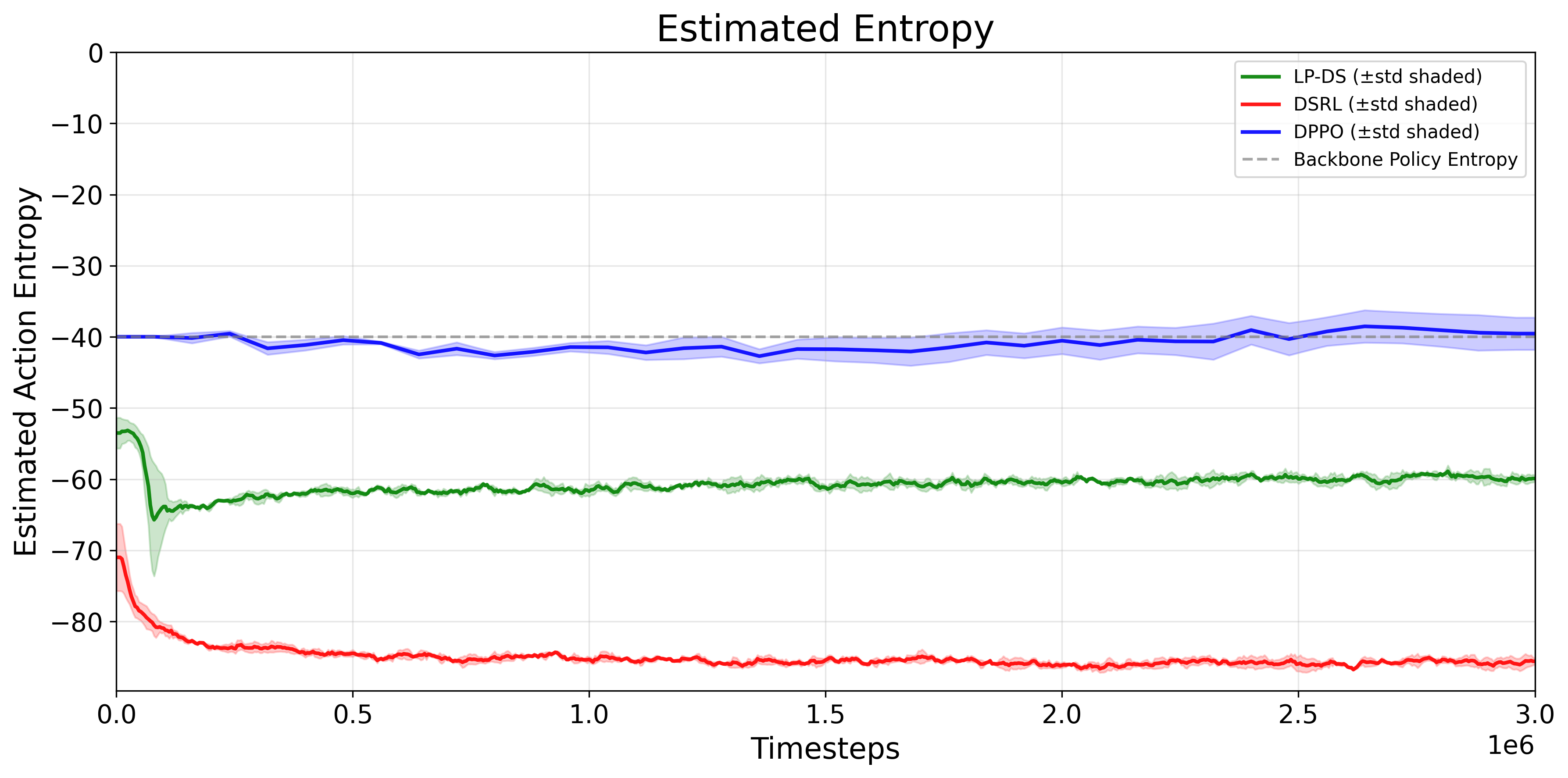}
    \caption{\textbf{Estimated action entropy during online adaptation (\textsc{Adroit Pen}).}
    Results are obtained using the same training and evaluation configuration as the corresponding benchmark runs.
    We track Kozachenko--Leonenko k-NN action entropy while fine-tuning the same frozen backbone.
    Noise-space steering methods (LP-DS, DSRL) exhibit an inherent reduction in entropy relative to the backbone,
    whereas DPPO remains close to the backbone entropy.
    LP-DS consistently preserves higher entropy than DSRL.}
    \label{fig:entropy_comparison}
\end{figure}

Figure~\ref{fig:entropy_comparison} reports the estimated k-NN action entropy throughout online adaptation, averaged over 3 random seeds. The entropy is computed using the Kozachenko--Leonenko estimator described in Appendix~\ref{app:entropy_estimator}. Noise-space steering methods (\textbf{LP-DS} and \textbf{DSRL}) reduce entropy relative to the frozen backbone, as improvement is achieved by concentrating probability mass on higher-value regions of the latent noise space. This concentration translates into reduced diversity after decoding and can be interpreted as latent-space mode selection.

Importantly, \textbf{LP-DS consistently preserves higher action entropy than DSRL} while still improving performance (cf.~Figure~\ref{fig:baseline_comparison_all}).
This gap indicates that the Lagrangian trust-region mechanism in LP-DS mitigates premature collapse by discouraging overly aggressive concentration in the steering space.
In contrast, \textbf{DPPO} directly fine-tunes the policy in action space and therefore does not exhibit the same systematic entropy decay, remaining closer to the backbone entropy throughout training.

Together with the toy-domain analysis, these results show that LP-DS reduces mode collapse and better utilizes the multimodal capacity of the frozen generative backbone while achieving strong task performance.

\subsection{Trust-Region Ablations}
\label{sec:ablation}
\begin{figure}[t]
    \centering
    \includegraphics[width=\columnwidth]{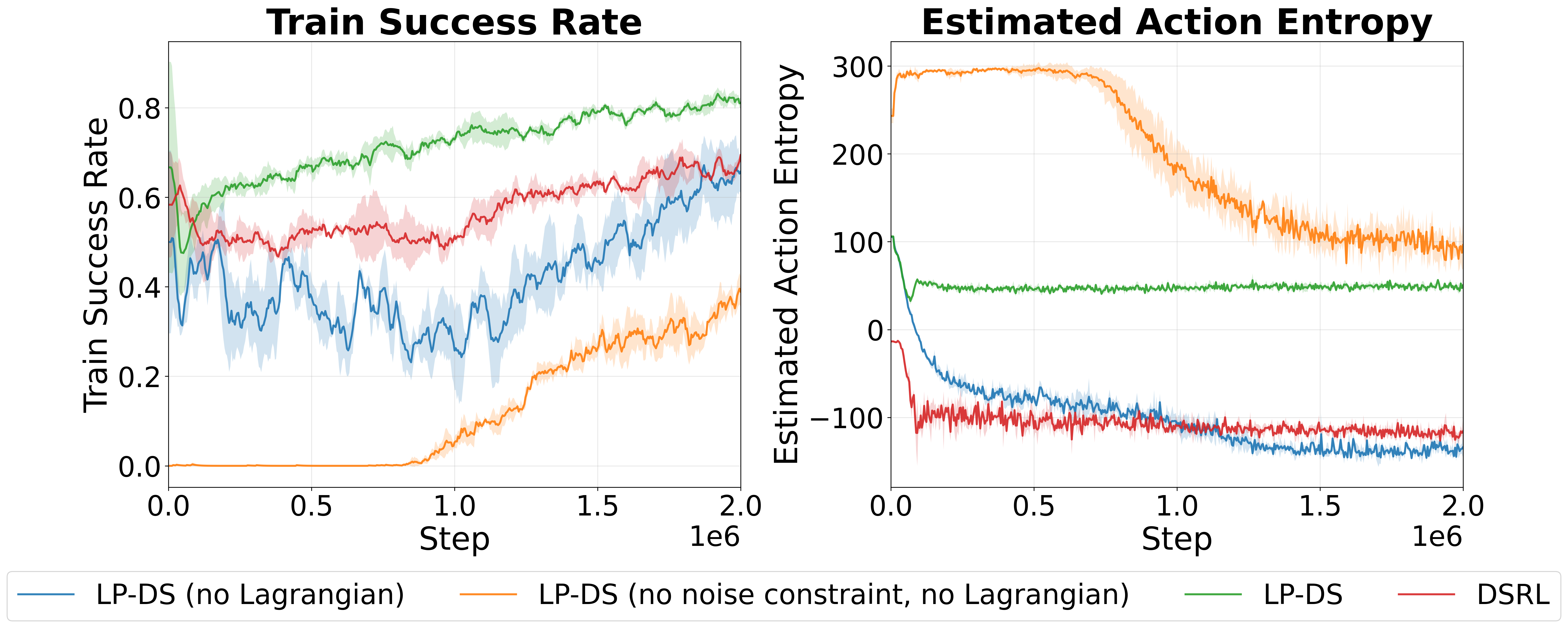}
    \caption{\textbf{Trust-region ablations on \textsc{Adroit Pen}.}
    We plot training success rate (EMA) and Kozachenko--Leonenko $k$-NN action entropy during online adaptation, using the same training and evaluation configuration as the corresponding benchmark runs and the same frozen backbone across methods.
    Removing the Lagrangian dual update destabilizes training and reduces final success; removing both the Lagrangian update and the noise bound leads to the worse performance.
    The full \textbf{LP-DS} objective attains the highest success while retaining higher action entropy than the ablated variants and DSRL.}
    \label{fig:pen_expert_metrics}
\end{figure}

We study the contribution of the trust-region components in LP-DS by ablating
(i) the Lagrangian penalty on the perturbation magnitude and
(ii) the explicit bound on the steering noise.
Figure~\ref{fig:pen_expert_metrics} reports training success rate (EMA) and estimated action entropy on \textsc{Pen}, comparing
\textbf{LP-DS},
\textbf{LP-DS w/o Lagrangian},
\textbf{LP-DS w/o Lagrangian \& noise bound},
and \textbf{DSRL}.

Removing the Lagrangian penalty leads to unstable learning and degraded success, while removing the noise bound alone does not cause severe performance degradation.
As also observed in Figure~\ref{fig:noise_magnitude}, LP-DS without hard clipping remains competitive across several environments.
In contrast, removing both constraints results in highly unstable behavior:
\textbf{LP-DS w/o Lagrangian \& noise bound} frequently explores extreme regions of the steering space that are unlikely under the Gaussian prior used during pretraining, producing low success rates and unreliable learning dynamics.
These results indicate that while hard bounding can improve stability, the Lagrangian trust-region constraint is the primary mechanism preventing out-of-distribution steering and preserving meaningful decoding by the frozen generative backbone.

Removing only the \emph{Lagrangian control} yields a different failure mode.
\textbf{LP-DS w/o Lagrangian} achieves moderate success but exhibits a steady decrease in estimated action entropy over training.
This behavior indicates progressive collapse toward a narrow subset of decoded behaviors, consistent with the actor converging to a small region of the steering space (often near a single effective noise realization).
In contrast, \textbf{LP-DS} maintains substantially higher entropy throughout training while also achieving the strongest success rate, demonstrating that the Lagrangian trust-region objective provides a practical mechanism for simultaneously improving task performance and preserving multimodal behavior.

Finally, \textbf{DSRL} exhibits noticeably lower action entropy compared to LP-DS in this setting, reflecting a stronger tendency toward collapse under noise-space steering when deviations are not explicitly regulated.
Overall, these ablations support the design choice of adaptive Lagrangian penalty: the Lagrangian update discourages collapse by regulating how aggressively the steering mechanism departs from the base sampling distribution.

\subsection{Sensitivity to the Trust-Region Target}
\label{sec:delta_sensitivity}

We next study whether LP-DS requires careful tuning of the trust-region target $\delta$. We sweep $\delta$ across four representative environments: \textsc{Hopper}, \textsc{Walker2d}, \textsc{RoboMimic Square}, and \textsc{Adroit Relocate}. Across these environments, we observe that the method is not highly sensitive to the exact value of $\delta$: values above $0.1$ generally yield strong performance, and nearby choices such as $0.35$, $0.5$, and $0.66$ lead to similar reward or success trends. This suggests that $\delta$ acts primarily as a coarse behavioral dial rather than a brittle hyperparameter requiring fine-grained tuning.

Smaller values of $\delta$ enforce conservative steering and better preserve the frozen prior, while larger values permit more aggressive reward maximization. This behavior is consistent with the toy-domain analysis in Section~\ref{sec:toy_delta}, where $\delta$ controls the trade-off between multimodal preservation and task specialization. Full sweep curves are provided in Appendix~\ref{app:delta_sensitivity}.

\subsection{Action-Space vs. Noise-Space Perturbation}
\label{sec:action_vs_noise}

\begin{figure}[t]
    \centering
    \includegraphics[width=.8\columnwidth]
    {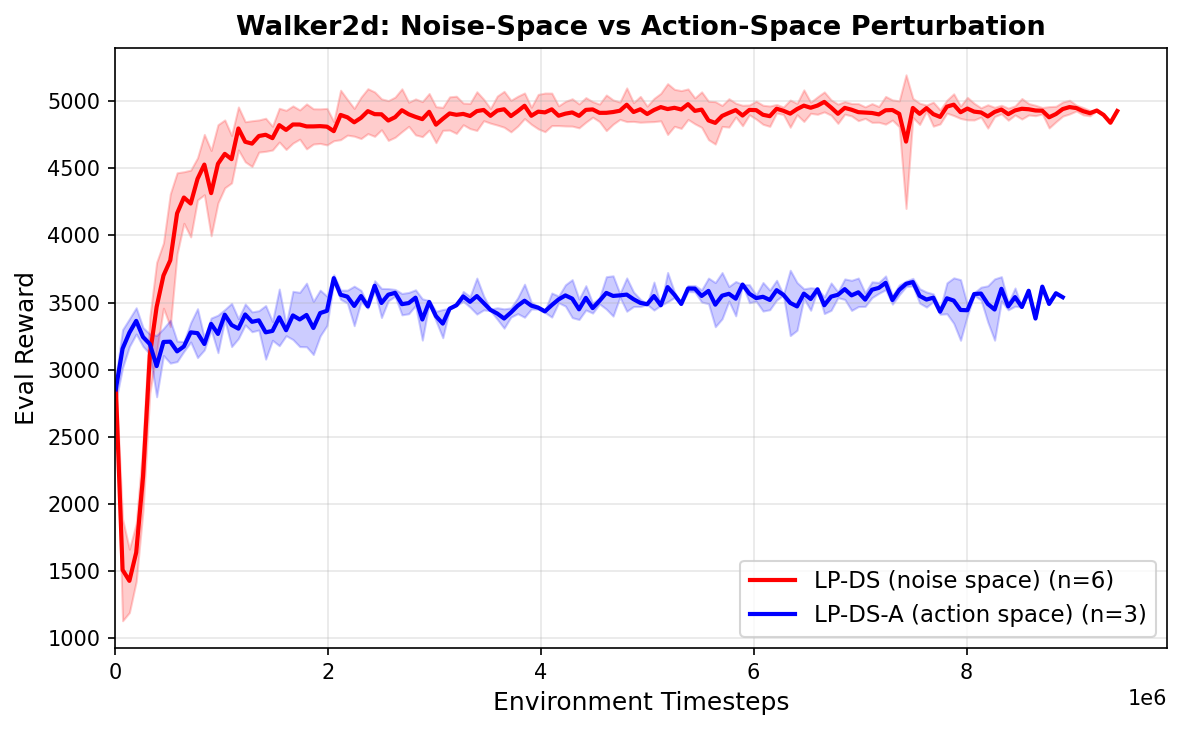}
    \caption{\textbf{Action-space vs. noise-space perturbation on \textsc{Walker2d}.}
    LP-DS optimizes a residual in the latent noise space using the latent critic $Q^{\mathcal W}$, while LP-DS-A applies the residual directly in action space using only the action critic $Q^{\mathcal A}$.
    Noise-space perturbation substantially outperforms direct action-space perturbation, indicating that steering the initial latent query is crucial for effectively adapting the frozen generative decoder.}
    \label{fig:walker_action_noise_ablation}
\end{figure}

To test whether optimizing in the latent noise space is necessary, we compare LP-DS against an action-space variant, denoted LP-DS-A, that applies the learned residual directly to the decoded action and optimizes it using the action critic $Q^{\mathcal A}$. This ablation removes the latent critic $Q^{\mathcal W}$ from the actor update and bypasses optimization over the initial generative noise. As shown in Figure~\ref{fig:walker_action_noise_ablation}, action-space perturbation provides only limited improvement and quickly plateaus around a substantially lower return, whereas noise-space LP-DS continues to improve and reaches much higher final performance. This result supports the role of the latent critic: for high-capacity generative decoders, optimizing the initial noise query is more effective than applying local corrections after decoding.

\subsection{Cross-Architecture Robustness}
\label{sec:cross_architecture}
We next evaluate whether LP-DS depends on a specific generative backbone. Beyond compact diffusion and flow-matching policies, we test LP-DS on the LIBERO-90 benchmark "pick up the cream cheese and put it in the tray" task \cite{liu2023liberobenchmarkingknowledgetransfer} using a large vision-language-action backbone. In this setting, the pretrained $\pi_0$ decoder~\citep{black2024pi0} is kept frozen and only a lightweight perturbation module is trained.

We also conduct a controlled comparison between diffusion and flow-matching backbones on \textsc{Hopper-v2} under matched settings. LP-DS achieves comparable final performance with both backbone classes, indicating that the residual perturbation and trust-region formulation are not tied to denoising diffusion chains. The full comparison is provided in Appendix~\ref{app:backbone_comparison}.

\begin{figure}[t]
    \centering
    \includegraphics[width=.8\columnwidth]{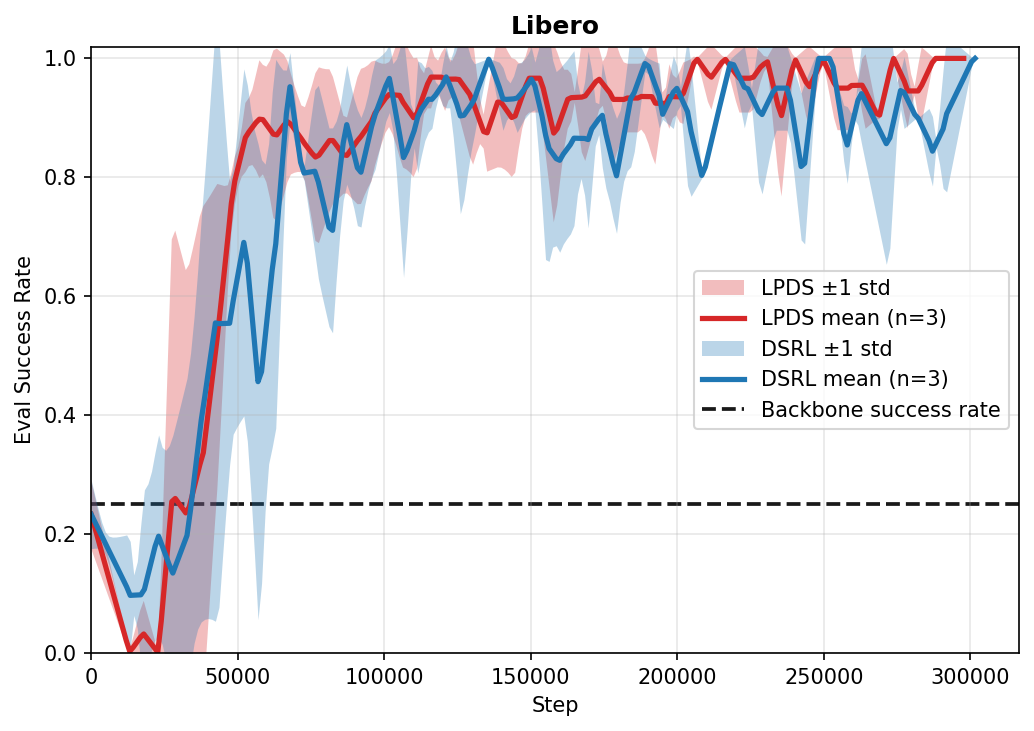}
    \caption{\textbf{LP-DS on LIBERO with a large VLA backbone.}
    LP-DS steers a frozen $\pi_0$ vision-language-action backbone using a lightweight perturbation module and improves substantially over the frozen-backbone success rate.
    The result demonstrates that LP-DS can scale beyond compact generative policies to large Transformer-based VLA models.}
    \label{fig:libero_pi0}
\end{figure}

\subsection{Real-World Robotic Deployment}
\label{sec:real_robot}

\begin{figure}[t]
    \centering
    \includegraphics[width=\columnwidth]{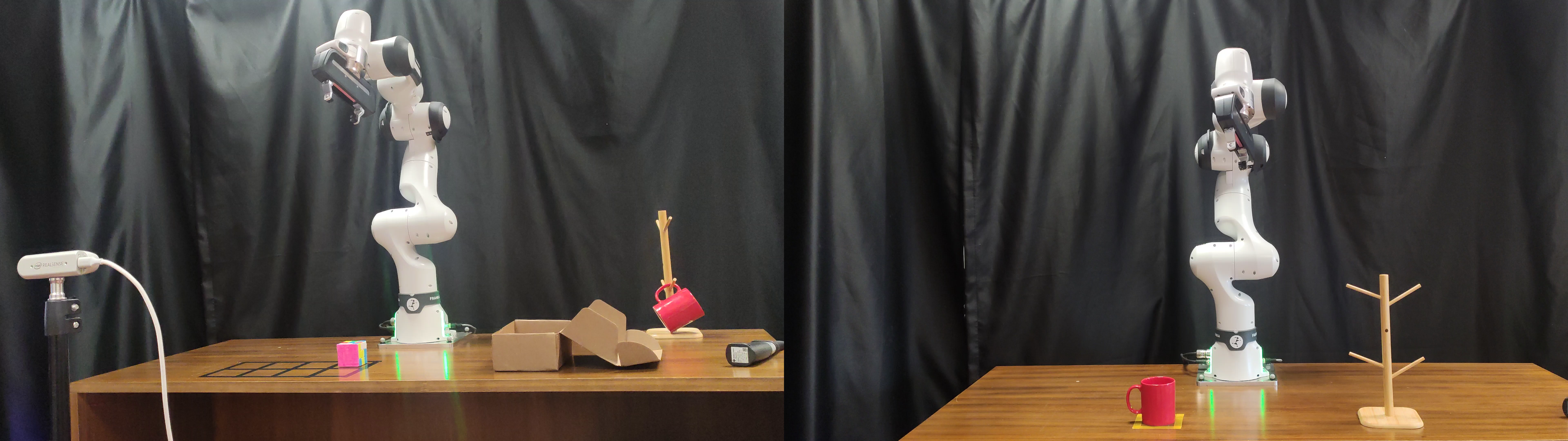}
    \caption{\textbf{Real-world Franka tasks.}
    Left: spatial pick-and-place setup, evaluated across a $2\times4$ grid of cube initial positions.
    Right: mug-hanging setup, where the robot must grasp the mug and align its handle with the wooden holder for insertion.}
    \label{fig:real_robot_tasks}
\end{figure}

We additionally evaluate LP-DS on a physical Franka Panda robot to test whether simulation-trained latent steering can transfer to real robot execution. We consider RGBD-conditioned spatial pick-and-place and high-precision mug hanging, shown in Figure~\ref{fig:real_robot_tasks}. In both tasks, the backbone policy is trained from human teleoperation demonstrations and kept frozen; LP-DS performs RL adaptation in a task-matched simulation environment, optimizing only the lightweight latent perturbation module before deploying the resulting steered policy on the physical robot. For pick-and-place, evaluated across a $2\times4$ grid with five trials per position, the frozen backbone succeeds in 18/40 trials, while LP-DS succeeds in 33/40 trials. For mug hanging, the frozen backbone succeeds in 11/20 trials, while LP-DS succeeds in 17/20 trials. These results provide initial physical-robot evidence that constrained latent steering can improve frozen generative policies beyond the simulation environment used for adaptation; details are provided in Appendix~\ref{app:real_robot}.

\section{Conclusion}
\label{sec:conclusion}

We presented \textbf{Lagrangian Perturbation Diffusion Steering (LP-DS)}, a lightweight online adaptation method for frozen generative policies. LP-DS improves performance by learning a state- and noise-conditioned residual perturbation of the latent prior, steering action generation without modifying the underlying diffusion or flow-matching decoder. The method uses a primal--dual Lagrangian trust-region objective to regulate perturbation magnitude and prevent off-prior latent queries.

Across RoboMimic, OpenAI Gym, Adroit, LIBERO, and physical Franka experiments, LP-DS improves success rates and returns over representative baselines while preserving more behavioral diversity than unconstrained noise-space steering. The trust-region target $\delta$ provides an interpretable control knob for balancing reward maximization and prior preservation. Future work includes adaptive trust-region targets, tighter links between latent constraints and trajectory-level objectives, and broader deployment in partially observable and large-scale vision-language-action settings.

\section*{Impact Statement}

This paper presents a method for improving pretrained generative control policies through constrained latent-space adaptation, with the goal of advancing reinforcement learning and generative modeling for robotics and continuous control. By enabling sample-efficient improvement while keeping the generative decoder frozen, LP-DS may reduce the cost of adapting large policies and help preserve the behavioral structure of pretrained robot-control models.

As with other learning-based control methods, unsafe deployment in physical systems could lead to unintended behavior. LP-DS partially mitigates this risk by constraining latent perturbations through a trust-region mechanism, but practical deployment still requires appropriate safety checks, monitoring, and task-specific validation. We do not foresee societal concerns beyond those commonly associated with reinforcement learning, robotics, and generative control.

\bibliography{references}

@misc{wagenmaker2025steeringdiffusionpolicylatent,
      title={Steering Your Diffusion Policy with Latent Space Reinforcement Learning}, 
      author={Andrew Wagenmaker and Mitsuhiko Nakamoto and Yunchu Zhang and Seohong Park and Waleed Yagoub and Anusha Nagabandi and Abhishek Gupta and Sergey Levine},
      year={2025},
      eprint={2506.15799},
      archivePrefix={arXiv},
      primaryClass={cs.RO},
      url={https://arxiv.org/abs/2506.15799}, 
}

@misc{ren2024diffusionpolicypolicyoptimization,
      title={Diffusion Policy Policy Optimization}, 
      author={Allen Z. Ren and Justin Lidard and Lars L. Ankile and Anthony Simeonov and Pulkit Agrawal and Anirudha Majumdar and Benjamin Burchfiel and Hongkai Dai and Max Simchowitz},
      year={2024},
      eprint={2409.00588},
      archivePrefix={arXiv},
      primaryClass={cs.RO},
      url={https://arxiv.org/abs/2409.00588}, 
}

@misc{hansenestruch2023idqlimplicitqlearningactorcritic,
      title={IDQL: Implicit Q-Learning as an Actor-Critic Method with Diffusion Policies}, 
      author={Philippe Hansen-Estruch and Ilya Kostrikov and Michael Janner and Jakub Grudzien Kuba and Sergey Levine},
      year={2023},
      eprint={2304.10573},
      archivePrefix={arXiv},
      primaryClass={cs.LG},
      url={https://arxiv.org/abs/2304.10573}, 
}

@misc{wang2023diffusionpoliciesexpressivepolicy,
      title={Diffusion Policies as an Expressive Policy Class for Offline Reinforcement Learning}, 
      author={Zhendong Wang and Jonathan J Hunt and Mingyuan Zhou},
      year={2023},
      eprint={2208.06193},
      archivePrefix={arXiv},
      primaryClass={cs.LG},
      url={https://arxiv.org/abs/2208.06193}, 
}

@misc{chi2024diffusionpolicyvisuomotorpolicy,
      title={Diffusion Policy: Visuomotor Policy Learning via Action Diffusion}, 
      author={Cheng Chi and Zhenjia Xu and Siyuan Feng and Eric Cousineau and Yilun Du and Benjamin Burchfiel and Russ Tedrake and Shuran Song},
      year={2024},
      eprint={2303.04137},
      archivePrefix={arXiv},
      primaryClass={cs.RO},
      url={https://arxiv.org/abs/2303.04137}, 
}

@misc{park2025flowqlearning,
      title={Flow Q-Learning}, 
      author={Seohong Park and Qiyang Li and Sergey Levine},
      year={2025},
      eprint={2502.02538},
      archivePrefix={arXiv},
      primaryClass={cs.LG},
      url={https://arxiv.org/abs/2502.02538}, 
}

@inproceedings{black2024pi0,
  title     = {{$\pi_0$}: A Vision-Language-Action Flow Model for General Robot Control},
  author    = {Black, Kevin and Brown, Noah and Driess, Danny and Esmail, Adnan and Equi, Michael and Finn, Chelsea and Fusai, Niccolo and Groom, Lachy and Hausman, Karol and Ichter, Brian and Jakubczak, Szymon and Jones, Tim and Ke, Liyiming and Levine, Sergey and Li-Bell, Adrian and Mothukuri, Mohith and Nair, Suraj and Pertsch, Karl and Shi, Lucy Xiaoyang and Tanner, James and Vuong, Quan and Walling, Anna and Wang, Haohuan and Zhilinsky, Ury},
  booktitle = {Proceedings of Robotics: Science and Systems (RSS)},
  year      = {2025}
}

@misc{ho2022imagenvideohighdefinition,
      title={Imagen Video: High Definition Video Generation with Diffusion Models}, 
      author={Jonathan Ho and William Chan and Chitwan Saharia and Jay Whang and Ruiqi Gao and Alexey Gritsenko and Diederik P. Kingma and Ben Poole and Mohammad Norouzi and David J. Fleet and Tim Salimans},
      year={2022},
      eprint={2210.02303},
      archivePrefix={arXiv},
      primaryClass={cs.CV},
      url={https://arxiv.org/abs/2210.02303}, 
}

@misc{mehta2024stablebc,
  title={Stable-BC: Controlling Covariate Shift with Stable Behavior Cloning},
  author={Mehta, Shaunak A. and Ciftci, Yusuf Umut and Ramachandran, Balamurugan and Bansal, Somil and Losey, Dylan P.},
  year={2024},
  eprint={2408.06246},
  archivePrefix={arXiv},
  primaryClass={cs.RO}
}

@book{sutton2018reinforcement,
  title     = {Reinforcement Learning: An Introduction},
  author    = {Sutton, Richard S. and Barto, Andrew G.},
  year      = {2018},
  edition   = {Second},
  publisher = {MIT Press},
  address   = {Cambridge, MA},
}

@misc{chandra2025diwadiffusionpolicyadaptation,
      title={DiWA: Diffusion Policy Adaptation with World Models}, 
      author={Akshay L Chandra and Iman Nematollahi and Chenguang Huang and Tim Welschehold and Wolfram Burgard and Abhinav Valada},
      year={2025},
      eprint={2508.03645},
      archivePrefix={arXiv},
      primaryClass={cs.RO},
      url={https://arxiv.org/abs/2508.03645}, 
}

@article{kozachenko1987sample,
  title   = {Sample Estimate of the Entropy of a Random Vector},
  author  = {Kozachenko, L. F. and Leonenko, N. N.},
  journal = {Problems of Information Transmission},
  volume  = {23},
  number  = {2},
  pages   = {95--101},
  year    = {1987}
}

@misc{pearce2023imitatinghumanbehaviourdiffusion,
      title={Imitating Human Behaviour with Diffusion Models}, 
      author={Tim Pearce and Tabish Rashid and Anssi Kanervisto and Dave Bignell and Mingfei Sun and Raluca Georgescu and Sergio Valcarcel Macua and Shan Zheng Tan and Ida Momennejad and Katja Hofmann and Sam Devlin},
      year={2023},
      eprint={2301.10677},
      archivePrefix={arXiv},
      primaryClass={cs.AI},
      url={https://arxiv.org/abs/2301.10677}, 
}

@misc{ze20243ddiffusionpolicygeneralizable,
      title={3D Diffusion Policy: Generalizable Visuomotor Policy Learning via Simple 3D Representations}, 
      author={Yanjie Ze and Gu Zhang and Kangning Zhang and Chenyuan Hu and Muhan Wang and Huazhe Xu},
      year={2024},
      eprint={2403.03954},
      archivePrefix={arXiv},
      primaryClass={cs.RO},
      url={https://arxiv.org/abs/2403.03954}, 
}

@misc{sridhar2023nomadgoalmaskeddiffusion,
      title={NoMaD: Goal Masked Diffusion Policies for Navigation and Exploration}, 
      author={Ajay Sridhar and Dhruv Shah and Catherine Glossop and Sergey Levine},
      year={2023},
      eprint={2310.07896},
      archivePrefix={arXiv},
      primaryClass={cs.RO},
      url={https://arxiv.org/abs/2310.07896}, 
}

@misc{dasari2024ingredientsroboticdiffusiontransformers,
      title={The Ingredients for Robotic Diffusion Transformers}, 
      author={Sudeep Dasari and Oier Mees and Sebastian Zhao and Mohan Kumar Srirama and Sergey Levine},
      year={2024},
      eprint={2410.10088},
      archivePrefix={arXiv},
      primaryClass={cs.RO},
      url={https://arxiv.org/abs/2410.10088}, 
}

@misc{nvidia2025gr00tn1openfoundation,
      title={GR00T N1: An Open Foundation Model for Generalist Humanoid Robots}, 
      author={NVIDIA and : and Johan Bjorck and Fernando Castañeda and Nikita Cherniadev and Xingye Da and Runyu Ding and Linxi "Jim" Fan and Yu Fang and Dieter Fox and Fengyuan Hu and Spencer Huang and Joel Jang and Zhenyu Jiang and Jan Kautz and Kaushil Kundalia and Lawrence Lao and Zhiqi Li and Zongyu Lin and Kevin Lin and Guilin Liu and Edith Llontop and Loic Magne and Ajay Mandlekar and Avnish Narayan and Soroush Nasiriany and Scott Reed and You Liang Tan and Guanzhi Wang and Zu Wang and Jing Wang and Qi Wang and Jiannan Xiang and Yuqi Xie and Yinzhen Xu and Zhenjia Xu and Seonghyeon Ye and Zhiding Yu and Ao Zhang and Hao Zhang and Yizhou Zhao and Ruijie Zheng and Yuke Zhu},
      year={2025},
      eprint={2503.14734},
      archivePrefix={arXiv},
      primaryClass={cs.RO},
      url={https://arxiv.org/abs/2503.14734}, 
}

@misc{kostrikov2021offlinereinforcementlearningimplicit,
      title={Offline Reinforcement Learning with Implicit Q-Learning}, 
      author={Ilya Kostrikov and Ashvin Nair and Sergey Levine},
      year={2021},
      eprint={2110.06169},
      archivePrefix={arXiv},
      primaryClass={cs.LG},
      url={https://arxiv.org/abs/2110.06169}, 
}

@misc{ajay2023conditionalgenerativemodelingneed,
      title={Is Conditional Generative Modeling all you need for Decision-Making?}, 
      author={Anurag Ajay and Yilun Du and Abhi Gupta and Joshua Tenenbaum and Tommi Jaakkola and Pulkit Agrawal},
      year={2023},
      eprint={2211.15657},
      archivePrefix={arXiv},
      primaryClass={cs.LG},
      url={https://arxiv.org/abs/2211.15657}, 
}

@misc{kang2023efficientdiffusionpoliciesoffline,
      title={Efficient Diffusion Policies for Offline Reinforcement Learning}, 
      author={Bingyi Kang and Xiao Ma and Chao Du and Tianyu Pang and Shuicheng Yan},
      year={2023},
      eprint={2305.20081},
      archivePrefix={arXiv},
      primaryClass={cs.LG},
      url={https://arxiv.org/abs/2305.20081}, 
}

@misc{chen2023offlinereinforcementlearninghighfidelity,
      title={Offline Reinforcement Learning via High-Fidelity Generative Behavior Modeling}, 
      author={Huayu Chen and Cheng Lu and Chengyang Ying and Hang Su and Jun Zhu},
      year={2023},
      eprint={2209.14548},
      archivePrefix={arXiv},
      primaryClass={cs.LG},
      url={https://arxiv.org/abs/2209.14548}, 
}

@misc{eyring2024renoenhancingonesteptexttoimage,
      title={ReNO: Enhancing One-step Text-to-Image Models through Reward-based Noise Optimization}, 
      author={Luca Eyring and Shyamgopal Karthik and Karsten Roth and Alexey Dosovitskiy and Zeynep Akata},
      year={2024},
      eprint={2406.04312},
      archivePrefix={arXiv},
      primaryClass={cs.CV},
      url={https://arxiv.org/abs/2406.04312}, 
}

@misc{mao2024lotterytickethypothesisdenoising,
      title={The Lottery Ticket Hypothesis in Denoising: Towards Semantic-Driven Initialization}, 
      author={Jiafeng Mao and Xueting Wang and Kiyoharu Aizawa},
      year={2024},
      eprint={2312.08872},
      archivePrefix={arXiv},
      primaryClass={cs.CV},
      url={https://arxiv.org/abs/2312.08872}, 
}

@misc{samuel2023generatingimagesrareconcepts,
      title={Generating images of rare concepts using pre-trained diffusion models}, 
      author={Dvir Samuel and Rami Ben-Ari and Simon Raviv and Nir Darshan and Gal Chechik},
      year={2023},
      eprint={2304.14530},
      archivePrefix={arXiv},
      primaryClass={cs.CV},
      url={https://arxiv.org/abs/2304.14530}, 
}

@misc{singh2020parrotdatadrivenbehavioralpriors,
      title={Parrot: Data-Driven Behavioral Priors for Reinforcement Learning}, 
      author={Avi Singh and Huihan Liu and Gaoyue Zhou and Albert Yu and Nicholas Rhinehart and Sergey Levine},
      year={2020},
      eprint={2011.10024},
      archivePrefix={arXiv},
      primaryClass={cs.LG},
      url={https://arxiv.org/abs/2011.10024}, 
}

@misc{venkatraman2025outsourceddiffusionsamplingefficient,
      title={Outsourced diffusion sampling: Efficient posterior inference in latent spaces of generative models}, 
      author={Siddarth Venkatraman and Mohsin Hasan and Minsu Kim and Luca Scimeca and Marcin Sendera and Yoshua Bengio and Glen Berseth and Nikolay Malkin},
      year={2025},
      eprint={2502.06999},
      archivePrefix={arXiv},
      primaryClass={cs.LG},
      url={https://arxiv.org/abs/2502.06999}, 
}

@misc{ankile2024juicerdataefficientimitationlearning,
      title={JUICER: Data-Efficient Imitation Learning for Robotic Assembly}, 
      author={Lars Ankile and Anthony Simeonov and Idan Shenfeld and Pulkit Agrawal},
      year={2024},
      eprint={2404.03729},
      archivePrefix={arXiv},
      primaryClass={cs.RO},
      url={https://arxiv.org/abs/2404.03729}, 
}

@misc{eyring2025noisehypernetworksamortizingtesttime,
      title={Noise Hypernetworks: Amortizing Test-Time Compute in Diffusion Models}, 
      author={Luca Eyring and Shyamgopal Karthik and Alexey Dosovitskiy and Nataniel Ruiz and Zeynep Akata},
      year={2025},
      eprint={2508.09968},
      archivePrefix={arXiv},
      primaryClass={cs.LG},
      url={https://arxiv.org/abs/2508.09968}, 
}

@article{ho2020denoising,
  title        = {Denoising Diffusion Probabilistic Models},
  author       = {Ho, Jonathan and Jain, Ajay and Abbeel, Pieter},
  journal      = {arXiv preprint arXiv:2006.11239},
  year         = {2020}
}

@misc{ahn2022icanisay,
      title={Do As I Can, Not As I Say: Grounding Language in Robotic Affordances}, 
      author={Michael Ahn and Anthony Brohan and Noah Brown and Yevgen Chebotar and Omar Cortes and Byron David and Chelsea Finn and Chuyuan Fu and Keerthana Gopalakrishnan and Karol Hausman and Alex Herzog and Daniel Ho and Jasmine Hsu and Julian Ibarz and Brian Ichter and Alex Irpan and Eric Jang and Rosario Jauregui Ruano and Kyle Jeffrey and Sally Jesmonth and Nikhil J Joshi and Ryan Julian and Dmitry Kalashnikov and Yuheng Kuang and Kuang-Huei Lee and Sergey Levine and Yao Lu and Linda Luu and Carolina Parada and Peter Pastor and Jornell Quiambao and Kanishka Rao and Jarek Rettinghouse and Diego Reyes and Pierre Sermanet and Nicolas Sievers and Clayton Tan and Alexander Toshev and Vincent Vanhoucke and Fei Xia and Ted Xiao and Peng Xu and Sichun Xu and Mengyuan Yan and Andy Zeng},
      year={2022},
      eprint={2204.01691},
      archivePrefix={arXiv},
      primaryClass={cs.RO},
      url={https://arxiv.org/abs/2204.01691}, 
}

@article{song2020denoising,
  title        = {Denoising Diffusion Implicit Models},
  author       = {Song, Jiaming and Meng, Chenlin and Ermon, Stefano},
  journal      = {arXiv preprint arXiv:2010.02502},
  year         = {2020},
  url          = {https://arxiv.org/abs/2010.02502}
}

@inproceedings{lipman2023flowmatching,
  title        = {Flow Matching for Generative Modeling},
  author       = {Lipman, Yaron and Chen, Ricky T. Q. and Ben-Hamu, Heli and Nickel, Maximilian and Le, Matt},
  booktitle    = {International Conference on Learning Representations (ICLR)},
  year         = {2023},
  url          = {https://arxiv.org/abs/2210.02747}
}

@article{liu2022rectified,
  title        = {Learning to Generate and Transfer Data with Rectified Flow},
  author       = {Liu, Xingchao and Gong, Chengyue and Liu, Qiang},
  journal      = {arXiv preprint arXiv:2209.03003},
  year         = {2022},
  url          = {https://arxiv.org/abs/2209.03003}
}

@inproceedings{robomimic2021,
  title={What Matters in Learning from Offline Human Demonstrations for Robot Manipulation},
  author={Ajay Mandlekar and Danfei Xu and Josiah Wong and Soroush Nasiriany and Chen Wang and Rohun Kulkarni and Li Fei-Fei and Silvio Savarese and Yuke Zhu and Roberto Mart\'{i}n-Mart\'{i}n},
  booktitle={Conference on Robot Learning (CoRL)},
  year={2021}
}

@misc{brockman2016openaigym,
      title={OpenAI Gym}, 
      author={Greg Brockman and Vicki Cheung and Ludwig Pettersson and Jonas Schneider and John Schulman and Jie Tang and Wojciech Zaremba},
      year={2016},
      eprint={1606.01540},
      archivePrefix={arXiv},
      primaryClass={cs.LG},
      url={https://arxiv.org/abs/1606.01540}, 
}

@article{rajeswaran2017learning,
  title={Learning complex dexterous manipulation with deep reinforcement learning and demonstrations},
  author={Rajeswaran, Aravind and Kumar, Vikash and Gupta, Abhishek and Vezzani, Giulia and Schulman, John and Todorov, Emanuel and Levine, Sergey},
  journal={arXiv preprint arXiv:1709.10087},
  year={2017}
}

@misc{toussaint2026robotic,
  author       = {Toussaint, Marc},
  title        = {robotic: Robotic Control Interface \& Manipulation Planning Library},
  year         = {2026},
  publisher    = {PyPI},
  journal      = {Python Package Index},
  version      = {0.3.6},
  howpublished = {\url{https://pypi.org/project/robotic/0.3.6/}},
  note         = {Accessed: 2026}
}

@misc{jia2024diversebehaviorsbenchmarkimitation,
      title={Towards Diverse Behaviors: A Benchmark for Imitation Learning with Human Demonstrations}, 
      author={Xiaogang Jia and Denis Blessing and Xinkai Jiang and Moritz Reuss and Atalay Donat and Rudolf Lioutikov and Gerhard Neumann},
      year={2024},
      eprint={2402.14606},
      archivePrefix={arXiv},
      primaryClass={cs.RO},
      url={https://arxiv.org/abs/2402.14606}, 
}

@misc{liu2023liberobenchmarkingknowledgetransfer,
      title={LIBERO: Benchmarking Knowledge Transfer for Lifelong Robot Learning}, 
      author={Bo Liu and Yifeng Zhu and Chongkai Gao and Yihao Feng and Qiang Liu and Yuke Zhu and Peter Stone},
      year={2023},
      eprint={2306.03310},
      archivePrefix={arXiv},
      primaryClass={cs.AI},
      url={https://arxiv.org/abs/2306.03310}, 
}
\bibliographystyle{icml2026}

\clearpage

\appendix
\section{Additional Experimental Details}
\label{app:experiment_details}

\subsection{Common Experimental Settings}
\label{app:common_settings}

For fair comparison, we adopt the same environments, evaluation protocols, network architectures, optimizer settings, and training schedules as the DSRL baseline.
Unless otherwise specified, all hyperparameters for RoboMimic and OpenAI Gym experiments are identical to those reported in Appendix C.1 of the DSRL paper~\citep{wagenmaker2025steeringdiffusionpolicylatent}. Implementation given by DSRL is used to run the experiments. 
This includes backbone architectures, critic designs, replay buffer sizes, update-to-data ratios, and evaluation cadence.

For the baselines DPPO, IDQL, DQL, we used the hyperparameters and implementation provided by \citep{ren2024diffusionpolicypolicyoptimization}. These parameters are for OpenAI Gym locomotion and Robomimic environments. Please refer to section E10 of the related paper for further details.

LP-DS introduces a single additional hyperparameter beyond DSRL: the trust-region target $\delta$, which controls the allowable magnitude of the latent perturbation.
All other settings are inherited unchanged from the DSRL configuration.

\subsection{Adroit Experiments}
\label{sec:adroit_hparams}
Table~\ref{tab:adroit_hparams_detailed} reports the environment-specific hyperparameters used in our Adroit experiments. These values are used for LP-DS and DSRL experiments. Critic gradient steps per update is represented by $Q^W$. We set target entropy of DSRL to 0 on all Adroit tasks.

To train new diffusion policies, we used the code provided by DPPO \cite{ren2024diffusionpolicypolicyoptimization}.  Flow matching policies are trained from scratch.

For DPPO, IDQL, DQL, we used the parameters specified in Table \ref{tab:adroit_baseline_hparams}.

The same action chunk size ($T_a=8$) and prediction horizon ($T_p=1$) were used in all runs.

Table \ref{tab:clip_val} shows the clipping values used in algorithms LP-DS and DSRL in adroit experiments.

\subsection{Trust-Region Targets Across Environments}
\label{app:delta_values}

LP-DS introduces a single additional hyperparameter relative to DSRL: the trust-region target $\delta$.
For each environment, $\delta$ is fixed throughout training and selected to balance performance and behavioral diversity, as analyzed in Section~\ref{sec:toy_delta}.

Table~\ref{tab:delta_across_envs} summarizes the trust-region targets used for all benchmark families.

\FloatBarrier
\begin{table}[t!]
\centering
\caption{Hyperparameters for Adroit dexterous manipulation experiments used for LP-DS and DSRL}
\label{tab:adroit_hparams_detailed}
\resizebox{\columnwidth}{!}{%
\begin{tabular}{lcccc}
\toprule
\textbf{Hyperparameter}
& \textsc{Pen/Hammer/Door/Relocate}
 \\
\midrule
Hidden size                          & 256\\
Gradient steps per update (UTD)      & 10\\
$Q^W$ update steps                   & 10 \\
Discount factor ($\gamma$)           & 0.99 \\
Action magnitude bound    & 2.5 \\
Initial environment steps            & 2{,}000 \\
Backbone sampling                    & ODE \\
Number of actor/critic layers & 3 \\
Number of critics & 2\\

\end{tabular}%
}
\end{table}
\FloatBarrier



\begin{table}[t]
\centering
\caption{Trust-region target $\delta$ used by LP-DS across environments.}
\label{tab:delta_across_envs}

\begin{tabular}{lc}
\toprule
\textbf{Environment} & \textbf{Trust-region target $\delta$} \\
\midrule
\textsc{RoboMimic Square}   & 0.35 \\
\textsc{RoboMimic Lift}     & 0.10 \\
\textsc{RoboMimic Can}      & 0.35 \\
\midrule
\textsc{Hopper-v2}          & 0.50 \\
\textsc{Walker2D-v2}        & 0.35 \\
\textsc{HalfCheetah-v2}     & 0.35 \\
\midrule
\textsc{Adroit Pen}         & 0.35 \\
\textsc{Adroit Hammer}      & 0.10 \\
\textsc{Adroit Door}        & 0.35 \\
\textsc{Adroit Relocate}    & 0.35 \\
\bottomrule
\end{tabular}
\end{table}

\begin{table}[t]
\centering
\caption{Hard clipping values ($b_W$) used in Adroit environments}
\label{tab:clip_val}

\begin{tabular}{lc}
\toprule
\textbf{Environment} & \textbf{Clipping value} \\
\midrule
\textsc{Adroit Pen}         & 2.5 \\
\textsc{Adroit Hammer}      & 2.5 \\
\textsc{Adroit Door}        & 2.5 \\
\textsc{Adroit Relocate}    & 2.5 \\
\bottomrule
\end{tabular}
\end{table}

\begin{table}[t]
\centering
\caption{Hyperparameter settings for DPPO, DQL and IDQL used in the Adroit environment.}
\label{tab:adroit_baseline_hparams}
\begin{tabular}{l l r}
\toprule
Method & Hyperparameter & Value \\
\midrule

\multirow{7}{*}{\textbf{DPPO}}
& $\gamma_{\text{denoise}}$ & 0.99 \\
& GAE $\lambda$             & 0.95 \\
& $N_{\theta}$              & 5 \\
& $N_{\phi}$                & 5 \\
& $\epsilon$                & 0.01 \\
& Batch size                & 50{,}000 \\
& $K'$                      & 10 \\
\midrule

\multirow{5}{*}{\textbf{DQL}}
& $\alpha_{\text{DQL}}$     & 1 \\
& $N_{\theta}$              & 16 \\
& $N_{\phi}$                & 16 \\
& Replay buffer size        & 1{,}000{,}000 \\
& Batch size                & 500 \\
\midrule

\multirow{5}{*}{\textbf{IDQL}}
& $M_{\text{IDQL}}$         & 20 \\
& $N_{\theta}$              & 128 \\
& $N_{\phi}$                & 128 \\
& Replay buffer size        & 1{,}000{,}000 \\
& Batch size                & 1{,}000 \\
\bottomrule
\end{tabular}
\end{table}

\subsection{Evaluation Protocol}
\label{app:evaluation}

All results are averaged over multiple random seeds (6 for main benchmarks, 3 for entropy analyses and ablations).
We use identical seeds across methods to reduce variance.
Evaluation rollouts are performed using deterministic decoding for diffusion policies (DDIM) and deterministic ODE integration for flow-matching backbones.

Performance is reported as success rate for sparse-reward tasks and episodic return for dense-reward tasks, following standard benchmark conventions.

Action chunking is considered while calculating the step counts so that for all implementations, one step corresponds to the execution of one action.

\section{Toy Experiment}
\label{sec:toy_example}

We design a symmetric multi-goal navigation task to isolate trajectory-level mode collapse under online adaptation. The environment is a 2D continuous-control MDP with state $s=(x,y)\in[-2,2]^2$ and action $a=(\Delta x,\Delta y)\in[-1,1]^2$. Transitions follow $s_{t+1}=\mathrm{clip}(s_t + 0.5\,\mathrm{clip}(a_t,-1,1),-2,2)$ with a horizon of 20 steps. The reward is shaped by the distance to the nearest goal, with a sparse bonus upon entering a goal region: $r_t=-\min_{g\in\mathcal{G}}\|s_t-g\|_2 + 10\,\mathbb{I}[\min_{g\in\mathcal{G}}\|s_t-g\|_2<0.2]$, where $\mathcal{G}=\{(\pm1,\pm1)\}$ are four corner goals.

\begin{figure}[t!]
    \centering
    \includegraphics[width=\columnwidth]{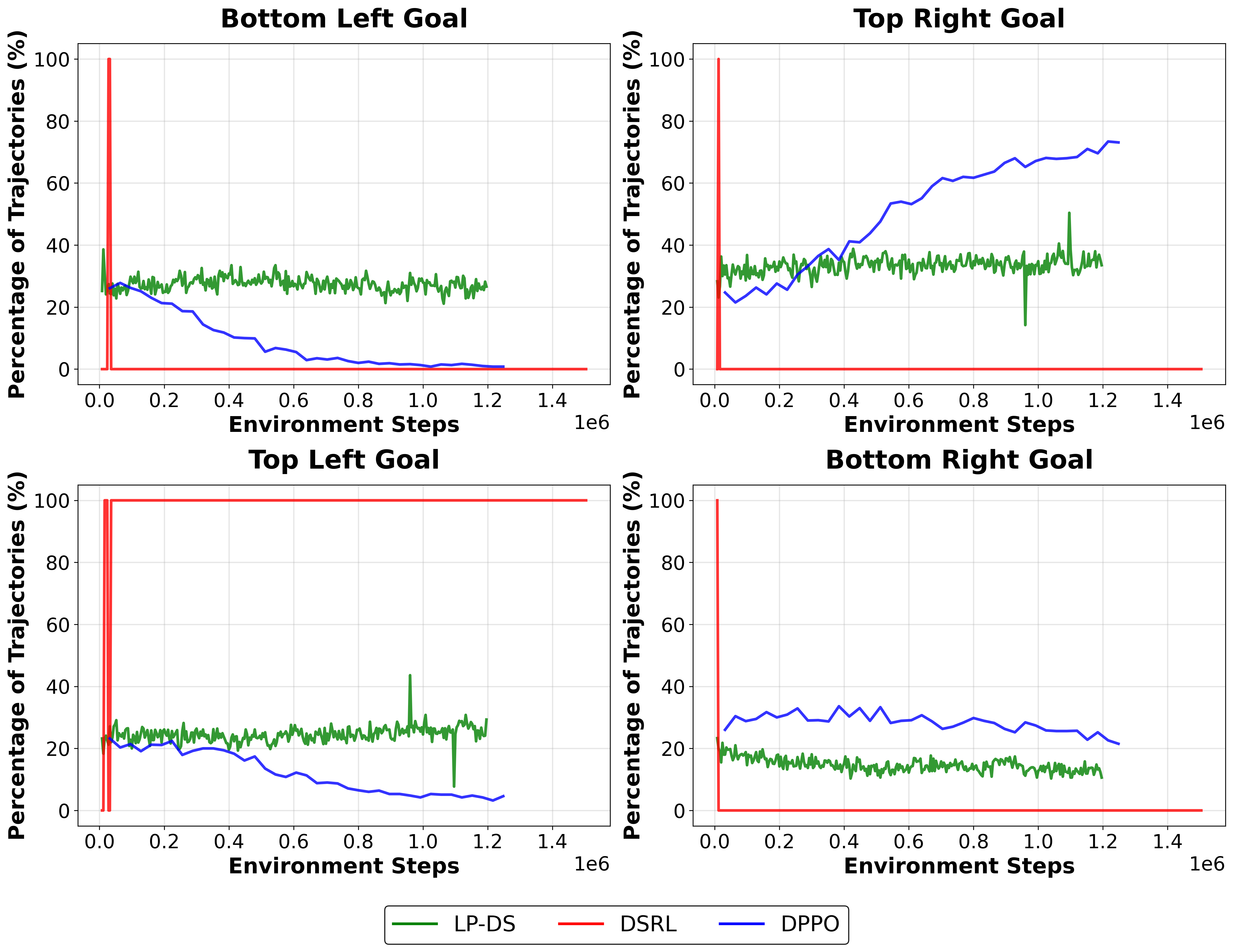}
    \caption{\textbf{Trajectory-level mode coverage in the symmetric multi-goal toy task.}
    Each panel corresponds to one goal (bottom-left, top-right, top-left, bottom-right) and shows, as training proceeds, the fraction of evaluation trajectories (out of 1000 rollouts per evaluation) that reach that goal.
    Concentration of mass into a single panel indicates mode collapse, while sustained non-trivial mass across multiple panels indicates preserved multimodality.
    }
    \label{fig:goal_succes}
\end{figure}

To obtain a multimodal behavioral prior, we generate an offline dataset of state--action pairs using a noisy expert rather than interacting with the environment. We sample states uniformly, $s\sim \mathrm{Unif}([-2,2]^2)$, choose the nearest goal $g^\star(s)=\arg\min_{g\in\mathcal{G}}\|s-g\|_2$, and define the ideal direction action $a^\star(s)=\mathrm{clip}(g^\star(s)-s,-1,1)$. To model imperfect demonstrations and induce stochasticity, we add i.i.d.\ Gaussian noise to the expert action and re-clip: $a=\mathrm{clip}(a^\star(s)+\xi,-1,1)$ with $\xi\sim\mathcal{N}(0,\sigma^2 I)$. Unless otherwise stated, we use $\sigma=0.1$ and generate $N=10^6$ samples, yielding a static dataset $\mathcal{D}_{\mathrm{off}}=\{(s_i,a_i)\}_{i=1}^{N}$.

We then train a diffusion policy backbone from scratch on $\mathcal{D}_{\mathrm{off}}$ via behavior cloning and freeze the decoder for all subsequent online adaptation methods. Starting from the same frozen backbone initialization, we adapt the policy using LP-DS, DSRL, and DPPO, and study whether adaptation preserves the backbone’s multi-goal coverage or collapses to a single goal.

Figure~\ref{fig:goal_succes} reports trajectory-level mode coverage during training. At each evaluation point, we run 1000 rollouts and assign each trajectory to one of the four goals based on the goal it reaches (e.g., the first goal region entered, or equivalently the closest goal at termination). Each subplot corresponds to one goal, and the curve value is the fraction of evaluation trajectories reaching that goal. LP-DS is run with parameter $\delta=.01$. A mode-preserving policy maintains non-trivial mass across multiple panels, while mode collapse appears as concentration of probability mass into a single panel. This visualization makes collapse directly observable at the trajectory level, complementing the action-space entropy analysis in Section~\ref{app:entropy_estimator}.

We used $b_W=1.5$ as the hard clipping bound for DSRL for this experiment.


\section{Kozachenko--Leonenko Action Entropy Estimator}
\label{app:entropy_estimator}

To quantify behavioral diversity in decoded action space, we estimate the conditional differential entropy of the policy-induced action distribution $p_\pi(a \mid s)$ using the Kozachenko--Leonenko $k$-nearest-neighbor estimator~\citep{kozachenko1987sample}. For probe states $\{s_b\}_{b=1}^B$ sampled from the replay buffer, we draw $K$ stochastic decoded action samples from the generative policy at each state, yielding $\{a_{b,i}\}_{i=1}^K$. Each action sample is flattened across the action-chunk dimensions before entropy estimation.

We compute the estimator independently for each probe state and then average across probe states:
\begin{align}
\widehat{H}_{\mathrm{KL}}(A \mid S)
&= \frac{1}{B}\sum_{b=1}^B 
\widehat{H}_{\mathrm{KL}}\big(\{a_{b,i}\}_{i=1}^K\big), \\
\widehat{H}_{\mathrm{KL}}\big(\{a_i\}_{i=1}^K\big)
&= -\psi(k) + \psi(K) + \log c_D \notag\\
&\quad + \frac{D}{K}\sum_{i=1}^K 
\log\!\big(\rho_i^{(k)} + \epsilon\big).
\end{align}
Here $D$ is the flattened action dimensionality, $c_D$ is the volume of the $D$-dimensional unit ball, $\psi(\cdot)$ is the digamma function, and $\rho_i^{(k)}$ is the Euclidean distance from sample $a_i$ to its $k$-th nearest neighbor. We add a small numerical constant $\epsilon$ for stability. A decrease in this estimate indicates that the decoded action distribution has become more concentrated around a smaller set of behaviors.


\section{Diffusion vs. Flow-Matching Backbone Comparison}
\label{app:backbone_comparison}

\begin{figure}[t]
    \centering
    \includegraphics[width=.8\columnwidth]{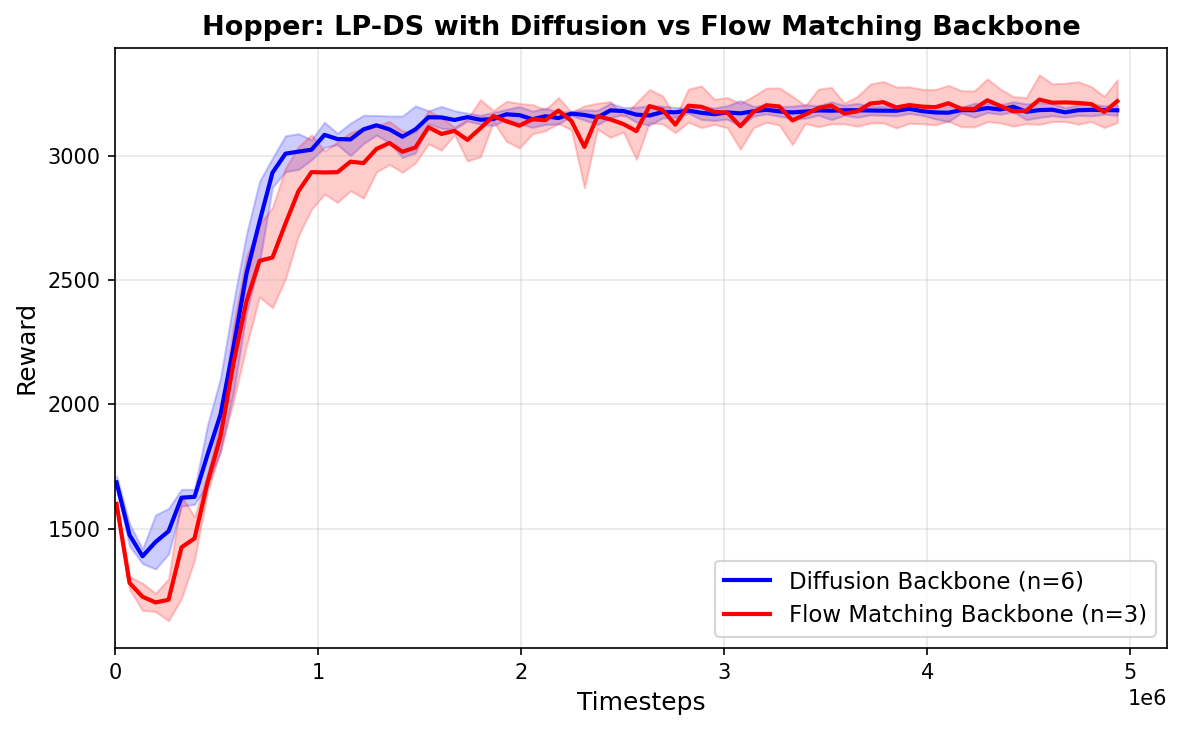}
    \caption{\textbf{Diffusion vs. flow-matching backbones on \textsc{Hopper-v2}.}
    LP-DS achieves comparable final performance when applied to either a diffusion backbone or a flow-matching backbone under matched settings.
    This indicates that the residual perturbation and trust-region formulation are not tied to denoising diffusion chains.}
    \label{fig:hopper_backbone_comparison}
\end{figure}

To isolate whether LP-DS depends on a particular generative-policy architecture, we compare diffusion and flow-matching backbones in the same \textsc{Hopper-v2} environment under matched settings. Figure~\ref{fig:hopper_backbone_comparison} shows that LP-DS achieves comparable final performance with both backbone classes. The flow-matching backbone initially learns slightly more slowly, but converges to the same performance range as the diffusion backbone. This supports the claim that LP-DS is not specific to denoising diffusion policies and can also steer ODE-based generative decoders.

\section{Trust-Region Target Sensitivity}
\label{app:delta_sensitivity}

\begin{figure*}[t!]
    \centering
    \includegraphics[width=\textwidth]{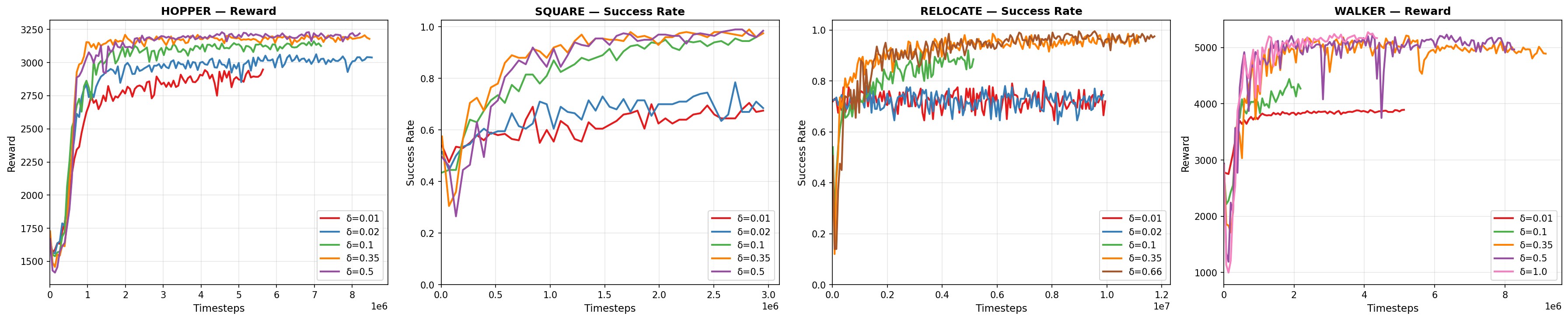}
    \caption{\textbf{Sensitivity to the trust-region target $\delta$.}
    We sweep $\delta$ across representative environments and report the resulting reward or success trends.
    Very small trust-region targets can overly restrict latent steering, while moderate values provide strong performance without requiring fine-grained tuning.
    The results indicate that $\delta$ acts as a coarse control knob for the trade-off between prior preservation and reward maximization.}
    \label{fig:delta_sensitivity}
\end{figure*}

LP-DS introduces the trust-region target $\delta$, which controls the allowable magnitude of the learned latent perturbation. In the main experiments, we use a fixed environment-specific value of $\delta$ as listed in Table~\ref{tab:delta_across_envs}. Here, we study how sensitive the method is to this choice by sweeping $\delta$ across representative environments and measuring the resulting reward or success trends.

Figure~\ref{fig:delta_sensitivity} shows the effect of different trust-region targets. Overall, LP-DS is not highly sensitive to the exact value of $\delta$ over a broad range. Very small values impose a conservative trust region, which strongly anchors the policy to the frozen backbone and can limit reward improvement. Larger values allow more aggressive latent-space steering and typically improve task performance, but excessively loose constraints may reduce the regularizing effect of the prior and increase the risk of mode collapse or unstable adaptation. Across the tested environments, moderate values such as $\delta=0.35$ provide a reliable default, while nearby values often yield similar performance.

These results support the interpretation of $\delta$ as a coarse behavioral dial rather than a brittle hyperparameter. Smaller values favor prior preservation and behavioral diversity, whereas larger values favor task specialization and reward maximization. This trend is consistent with the toy-domain analysis in Section~\ref{sec:toy_delta}, where increasing $\delta$ shifts LP-DS from conservative multimodal behavior toward more concentrated high-reward behavior.

\section{Avoiding Environment}
\label{app:avoiding}

\begin{figure}[t!]
    \centering
    \includegraphics[width=\columnwidth]{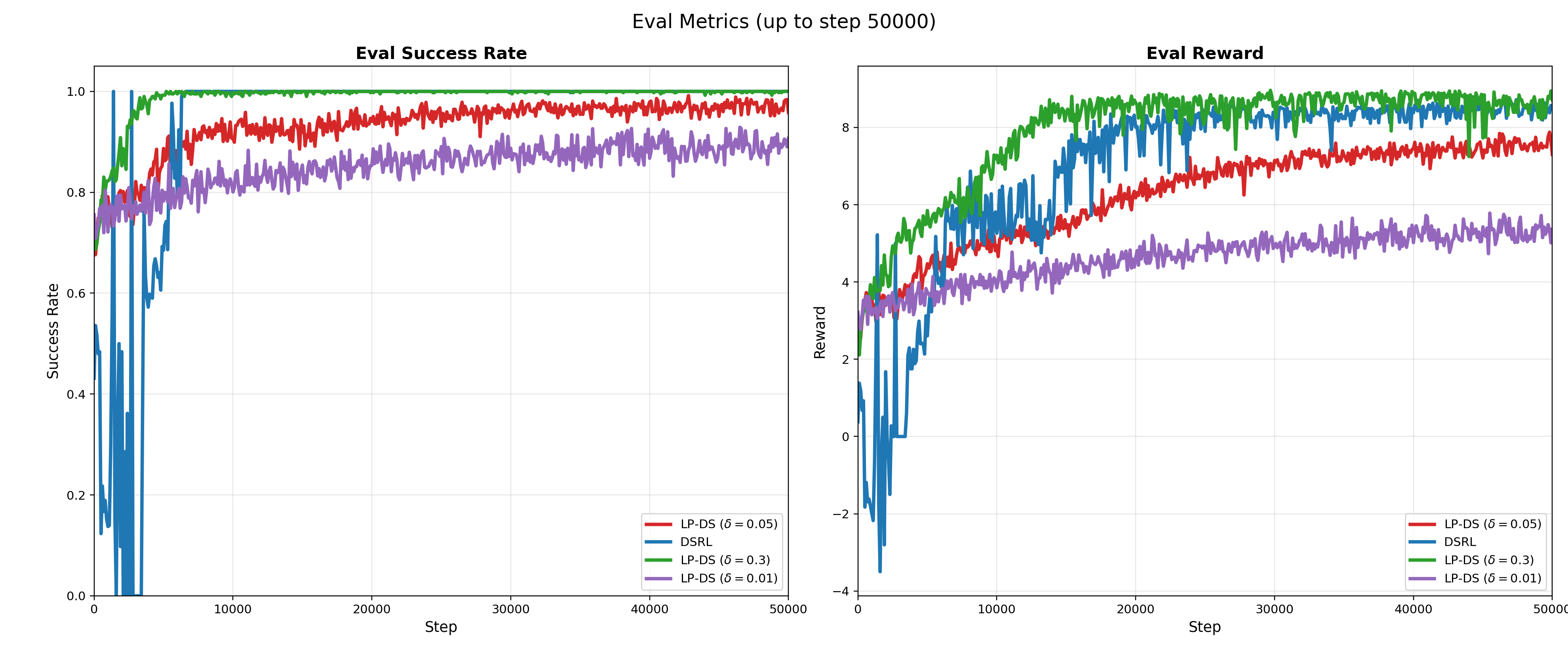}
    \caption{\textbf{Success rate and reward in the \textsc{Avoiding} environment.}
    We report evaluation success rate and evaluation reward for LP-DS with different trust-region targets and for DSRL. The results are consistent with the trajectory visualizations: smaller trust-region targets yield more conservative but diverse behavior, while larger targets produce stronger specialization and higher final task performance.}
    \label{fig:avoiding_metrics}
\end{figure}

Figure~\ref{fig:avoiding_metrics} reports quantitative success-rate and reward curves for the \textsc{Avoiding} environment from \citet{jia2024diversebehaviorsbenchmarkimitation}. These curves are consistent with the trajectory visualizations in Section~\ref{sec:avoiding_diversity}: larger trust-region targets improve specialization and final performance, while smaller targets preserve more diverse behavior.

\section{Real-World Robot Experiments}
\label{app:real_robot}

We evaluate LP-DS on a physical Franka Panda robot using frozen generative backbones trained from human teleoperation demonstrations. The goal of these experiments is to test whether latent perturbation steering learned in simulation can transfer to real robot execution under sensing noise, calibration error, and actuation uncertainty. We consider two physical manipulation tasks, shown in Figure~\ref{fig:real_robot_tasks}: spatial pick-and-place and mug hanging. In the spatial pick-and-place task, the robot must grasp a cube and place it at a target location. In the mug-hanging task, the robot must grasp a mug and hang it by its handle on a stationary holder. Mug hanging is more precision-sensitive because success requires accurate handle alignment and narrow-tolerance insertion.

For each task, we first train a generative backbone policy from human teleoperation demonstrations. We make the generative model condition on the robot joint state, observation received from Intel RealSense Depth Camera 435i (only for pick and place task), and the gripper position. The backbone maps the current observation to an action chunk through a flow-matching decoder. After behavior-cloning pretraining, this decoder is kept frozen throughout LP-DS adaptation and physical deployment. For each physical task, we construct a corresponding task environment in the RAI/Robotic simulation framework~\citep{toussaint2026robotic}. LP-DS adaptation is then performed in this simulated environment rather than directly on the physical robot. During adaptation, only the lightweight latent perturbation module is optimized; the generative decoder remains fixed. The resulting steered policy is then deployed on the physical Franka robot without further fine-tuning of the decoder.

For the spatial pick-and-place task, we evaluate the deployed policy across a $2\times4$ grid of object initial positions in the physical robot workspace. Each grid location is evaluated with five independent trials, resulting in 40 total physical trials. Table~\ref{tab:real_robot_grid_success} reports the per-position success rates before and after LP-DS adaptation, using the first five trials from each grid location. Position 1 corresponds to the top-right cell of the workspace grid and position 8 corresponds to the bottom-left cell. The frozen backbone succeeds in 18/40 trials, while LP-DS succeeds in 33/40 trials. For the mug-hanging task, we evaluate 20 independent physical trials. The frozen backbone succeeds in 11/20 trials, while LP-DS succeeds in 17/20 trials. Overall physical robot results are summarized in Table~\ref{tab:real_robot_summary}.

\begin{table}[t]
\centering
\caption{\textbf{Per-position success rates for real-world spatial pick-and-place.}
Each cell corresponds to one physical object initial position in the $2\times4$ workspace grid and reports frozen-backbone success versus LP-DS success over five trials. The table layout matches the physical grid: the top row is positions $7,5,3,1$ and the bottom row is positions $8,6,4,2$.}
\label{tab:real_robot_grid_success}
\begin{tabular}{cccc}
\toprule
\textbf{Pos. 7} & \textbf{Pos. 5} & \textbf{Pos. 3} & \textbf{Pos. 1} \\
Frozen: 0/5 & Frozen: 3/5 & Frozen: 3/5 & Frozen: 1/5 \\
LP-DS: 3/5 & LP-DS: 4/5 & LP-DS: 4/5 & LP-DS: 4/5 \\
\midrule
\textbf{Pos. 8} & \textbf{Pos. 6} & \textbf{Pos. 4} & \textbf{Pos. 2} \\
Frozen: 1/5 & Frozen: 3/5 & Frozen: 4/5 & Frozen: 3/5 \\
LP-DS: 4/5 & LP-DS: 5/5 & LP-DS: 4/5 & LP-DS: 5/5 \\
\bottomrule
\end{tabular}
\end{table}

\begin{table}[t]
\centering
\caption{\textbf{Real-world Franka evaluation.}
LP-DS improves the frozen generative backbone after simulation-based latent adaptation and physical deployment.}
\label{tab:real_robot_summary}
\begin{tabular}{lcc}
\toprule
Task & Frozen backbone & LP-DS \\
\midrule
Spatial pick-and-place & 18/40 & 33/40 \\
Mug hanging & 11/20 & 17/20 \\
\bottomrule
\end{tabular}
\end{table}

These results provide initial evidence that simulation-trained latent steering can improve frozen generative policies when transferred to real robot execution. Since LP-DS modifies only the latent input to the frozen decoder, the deployed policy remains anchored to the pretrained behavioral prior while still benefiting from task-directed adaptation. Failures are mainly caused by grasp misalignment, visual localization error, or small execution errors near contact-rich parts of the task. Videos are provided on the project page.

\end{document}